\documentclass[twocolumn]{aastex63}
\usepackage{amsmath}

\usepackage{color}
\usepackage{listings}
\usepackage{verbatimbox}
\usepackage{scrextend}
\usepackage{harpoon}
\usepackage{esvect}

\newcommand{\vecc}[1]{\boldsymbol{\vec{#1}}}

\accepted{December 5, 2020}
\submitjournal{AJ}

\shorttitle{SDSS-V: Kaiju}
\shortauthors{Sayres et al.}
\graphicspath{{./}{figures/}}

\begin{document}

\title{SDSS-V Algorithms: Fast, Collision-Free Trajectory Planning for Heavily Overlapping Robotic Fiber Positioners}

\author{Conor Sayres}
\affiliation{Department of Astronomy, Box 351580, University of Washington, Seattle, WA 98195, USA}

\author{Jos{\'e} R. S{\'a}nchez-Gallego}
\affiliation{Department of Astronomy, Box 351580, University of Washington, Seattle, WA 98195, USA}

\author{Michael R. Blanton}
\affiliation{Center for Cosmology and Particle Physics, Department of Physics, New York University, 726 Broadway, New York, NY 10003, USA}

\author{Ricardo Araujo}
\affiliation{School of Engineering, Swiss Federal Institute of Technology in Lausanne (EPFL), 1015 Lausanne, Switzerland}

\author{Mohamed Bouri}
\affiliation{School of Engineering, Swiss Federal Institute of Technology in Lausanne (EPFL), 1015 Lausanne, Switzerland}

\author{Lo{\"i}c Grossen}
\affiliation{School of Engineering, Swiss Federal Institute of Technology in Lausanne (EPFL), 1015 Lausanne, Switzerland}

\author{Jean-Paul Kneib}
\affiliation{Institute of Physics, Laboratory of Astrophysics, {\'E}cole Polytechnique F{\'e}d{\'e}rale de Lausanne (EPFL), Observatoire de Sauverny, 1290 Versoix, Switzerland}

\author{Juna A. Kollmeier}
\affiliation{Observatories of the Carnegie Institution for Science, 813 Santa Barbara Street, Pasadena, CA 91101, USA}

\author{Luzius Kronig}
\affiliation{School of Engineering, Swiss Federal Institute of Technology in Lausanne (EPFL), 1015 Lausanne, Switzerland}

\author{Richard W. Pogge}
\affiliation{Department of Astronomy, Ohio State University, 140 West 18th Avenue, Columbus, OH 43210-1173}

\author{Sarah Tuttle}
\affiliation{Department of Astronomy, Box 351580, University of Washington, Seattle, WA 98195, USA}




\begin{abstract}

Robotic fiber positioner (RFP) arrays are becoming heavily adopted in wide field massively multiplexed spectroscopic survey instruments.  RFP arrays decrease nightly operational overheads through rapid reconfiguration between fields and exposures.  In comparison to similar instruments, SDSS-V has selected a very dense RFP packing scheme where any point in a field is typically accessible to three or more robots.  This design provides flexibility in target assignment.  However, the task of collision-less trajectory planning is especially challenging.  We present two multi-agent distributed control strategies that are highly efficient and computationally inexpensive for determining collision-free paths for RFPs in heavily overlapping workspaces.  We demonstrate that a reconfiguration path between two arbitrary robot configurations can be efficiently found if ``folded'' state, in which all robot arms are retracted and aligned in a lattice-like orientation, is inserted between the initial and final states.  Although developed for SDSS-V, the approach we describe is generic and so applicable to a wide range of RFP designs and layouts.  Robotic fiber positioner technology continues to advance rapidly, and in the near future ultra-densely packed RFP designs may be feasible.  Our algorithms are especially capable in routing paths in very crowded environments, where we see efficient results even in regimes significantly more crowded than the SDSS-V RFP design.

\end{abstract}



\section{Introduction} \label{sec:intro}


Robotic fiber positioners (RFPs) are emerging as a promising technology in the today's landscape of wide-field multi-object spectroscopic instruments and surveys.  Many projects (e.g. LAMOST \citealt{lamost}, PFS \citealt{pfs}, DESI \citealt{desi}, MOONS \citealt{moons}, 4MOST \citealt{4most}, SDSS-V \citealt{2017arXiv171103234K}) have adopted densely packed positioner arrays patrolling the telescope's focal plane to obtain massively multiplexed spectroscopic observations of hundreds to thousands of objects in a field.  These arrays minimize operational overhead through rapid reconfigurations between fields and exposures.  Generally, each robotic fiber positioner will patrol a relatively small circular zone of the focal plane.  To obtain complete focal plane coverage, RFP patrol zones or workspaces necessarily overlap.  A system in which RFPs may physically interfere must therefore have a motion planning strategy ensuring that robots do not collide, wedge, or deadlock during reconfiguration.

The difficulty of the reconfiguration problem scales with the number of positioners able to physically occupy the same workspace.  Both SDSS-V and MOONS have chosen densely packed fiber positioner layouts that exhibit a large amount of workspace overlap when compared to similar instruments.  For example, a point in the DESI focal plane will typically be accessible to only a single positioner, whereas any point in the SDSS-V focal plane will typically be accessible to 3 or 4 positioners.  A heavily shared workspace increases flexibility in target assignment at the cost of a heightened chance of collision during reconfiguration.  The challenge posed by RFP reconfiguration is essentially the ``Cocktail Party Problem" \citep{cocktail}, in which mobile agents must navigate around each other in a crowded environment to reach a destination; an analogous situation to the trajectory planning problem humans subconsciously solve when moving around at a crowded cocktail party.

The Cocktail Party Problem lies in the general field of distributed multi-agent coordination and control, sometimes called swarm control.  \cite{multiagentoverview} and \cite{taxonomy} provide comprehensive overviews of this rapidly advancing subject.  The swarm control objective is to model how collective behavior (e.g. pattern formation, motion planning, synchronization) emerges from networks of individual agents following relatively simple protocols.  As the number of agents in a network increases, a global (or centralized) optimization strategy becomes computationally prohibitive.  Distributed control strategies often alleviate this computational burden by delegating the decision-making to the agents themselves, who then act according to the state of their immediate environments.  Distributed control is well-suited for real-time optimization and decision making in large networks, and scales well as additional agents are added.  Applications in the field are typically focused on coordinated robotics problems like remote sensor and surveillance networks or unmanned aerial vehicle coordination.  The algorithms are diverse with strategies ranging from game-theory approaches \citep{gametheory}, to Markov-based processes \citep{mehran}, to biologically-inspired artificial potential function (APF) models \citep{flock}.

\cite{2014A&A...566A..84M} approached the RFP navigation challenge using control laws based on artificial potential functions.  For array layouts with slightly overlapping fiber patrol zones (e.g. DESI or PFS), this strategy demonstrates the successful convergence of robots to targets.  However, this algorithm struggles when directly applied to heavily overlapping RFP layouts like those of SDSS-V: large fractions of positioners experience a deadlocked state, where progress toward the target halts due to the inability of combinations of positioners to move past one another.  \cite{2018SPIE10702E..8KT} injected additional control layers to the existing algorithm to detect and mitigate deadlock situations.  Using this augmented APF algorithm they report convergence increases from $\sim$65\% to $\sim$80\% for SDSS-V geometries.  Recently, \cite{matin} further improved the APF strategy with the introduction of cooperative fields, in which they demonstrate completely successful path solutions for SDSS-V positioner geometries in grids of various sizes in a set of 12 simulations.

In this work we present an alternative and simplified approach to the RFP navigation problem with a pair of closely related algorithms that are directly applicable to the majority of existing RFP systems today.  The first algorithm is a greedy heuristic, and the second algorithm is a Markov chain variant of the greedy heuristic.  Collecting statistics from thousands of simulations, we measure the efficiency of each algorithm as the fraction of targets assigned to targets acquired under collision-avoidance constraints.  In parameter spaces relevant to SDSS-V, we find the efficiency of the greedy heuristic to be $\sim 99.2 \%$ and efficiency of the Markov variant to be $> 99.9 \%$.  The high efficiencies of these routines are largely attributable to a reverse path generation strategy, rather than overall algorithmic complexity.  We discuss tradeoffs between operational overheads and overall targeting efficiency in the context of each algorithm.  We explore the limits of each algorithm by scaling up the relative size of the robots within the array to simulate overly crowded environments.  The results suggest that high efficiency paths can be found for positioner arrays that are significantly more crowded than the SDSS-V positioner array.  This work provided important feedback for the SDSS-V positioner design, so these algorithms serve as useful tools for vetting the feasibility of current and future RFP instrument designs.

This paper is organized as follows.  In Section \ref{sec:sdss-v}, we provide an overview of the SDSS-V survey for which the algorithms presented here have been developed.  In Section \ref{sec:layout}, we present a layout describing the relevant geometries and kinematics of the robotic fiber positioners.  In Section \ref{sec:algorithm} we present the generic algorithms followed by a detailed analysis in Section \ref{sec:analysis}.  In Section \ref{sec:deployment} we discuss methods for the effective deployment of these algorithms in the presence realistic hardware constraints.  Finally, in \ref{sec:discussion} we discuss these algorithms in context of overall survey optimization and comment on the general relevance of this work.  Pseudo code for all routines is provided in the appendix, and SDSS-V's Kaiju\footnote{\url{https://github.com/sdss/kaiju}} package contains source code for the algorithms we present.

\pagebreak 
\section{SDSS-V Survey Overview} \label{sec:sdss-v}

The Sloan Digital Sky Survey (\citealt{York_2000}; SDSS I/II) is currently preparing for a fifth phase of operations.  For nearly two decades, SDSS has continuously served high quality, well documented, and accessible data.  The project has evolved over the years, beginning with SDSS I/II, which provided multiband photometry from a large camera that now resides at the Smithsonian \citep{Gunn_1998} and multi-object optical spectroscopy \citep{Smee_2013} using plug plates at Apache Point Observatory (APO).  SDSS-III \citep{2011AJ....142...72E} included new instrumentation at APO for conducting a near-infrared multi-object spectroscopic survey APOGEE \citep{2017AJ....154...94M}, a radial velocity planet finding survey MARVELS \citep{10.1117/12.826651}, and increased total fiber capacity of the optical spectographs.  SDSS-IV \citep{2017AJ....154...28B} saw the inclusion of the MaNGA integral field unit (IFU) survey of nearby galaxies \citep{2015ApJ...798....7B}, and built out infrastructure and instrumentation to extend the APOGEE survey to the southern hemisphere at Las Campanas Observatory (LCO).

SDSS-V \citep{2017arXiv171103234K} is an all-sky, multi-epoch spectroscopic mapping survey, operating for five years in both hemispheres.  SDSS-V has three survey components: the Local Volume Mapper (LVM), the Black Hole Mapper (BHM), and the Milky Way Mapper (MWM).  The LVM is an optical spectroscopic IFU survey using small telescopes that will obtain contiguous coverage over wide regions of the Milky Way and the Magellanic Clouds; it will not use a robotic fiber positioner array.  The BHM and MWM surveys will operate in tandem, sharing a common robotic focal plane system (FPS).  One FPS will be installed at the 2.5 meter Sloan Foundation Telescope at APO \citep{2006AJ....131.2332G}, and another FPS will be installed at the 2.5 meter du Pont Telescope at LCO \citep{1973ApOpt..12.1430B}.  The MWM and BHM surveys have already begun gathering early data at APO using plug plates.  The FPS systems will come online mid-2021.

The BHM is a very-wide area dual hemisphere survey to obtain multi-epoch optical spectroscopy of accreting black holes.  It will provide comprehensive follow-up of X-ray sources from the recently launched eROSITA space telescope \citep{2010SPIE.7732E..0UP}.  With its multi-epoch capability, BHM will provide black hole masses for supermassive black holes through the reverberation mapping technique \citep{2015ApJS..216....4S} as well as probe the dynamics of the black hole environment by obtaining multi-epoch spectroscopy for quasars and AGN with sensitivity to variability on timescales ranging from decades to months when combined with legacy SDSS datasets.

The MWM science program is an all-sky survey targeting over 6 million stars in the Milky Way for multi-epoch optical and near-infrared spectroscopy.  The survey will measure fundamental stellar characteristics (e.g. age, chemistry, kinematics),  yielding a dataset well-suited for investigating the history and evolution of the Galaxy's structure.  The MWM aims to double the number of spectroscopically observed objects in SDSS in a span of five years.  This target volume increase would be impossible to meet using the existing SDSS plug plate system, which requires approximately 20 minutes of operational overhead between fields.  The combined MWM and BHM target volume drives the need for a nimble robotic fiber positioning system in SDSS-V.

To carry out the BHM and MWM surveys, a pair of SDSS-V robotic fiber positioning systems are under construction.  The FPS at APO will cover a 7 deg$^2$ field of view.  The FPS at LCO will cover a 3 deg$^2$ field of view.  Each FPS consists of a hexagonally-packed grid of 547 positions.  Robotic fiber positioners will occupy only 500 of these positions.  The remainder of the positions are populated with low-profile fiducial elements for a visual feedback system.  Each robot has the capacity to carry two science fibers in a common ferrule.  All 500 positioners will feed a BOSS optical spectrograph \citep{2013AJ....146...32S}, while a subset of up to 300 robots will simultaneously feed an APOGEE near-infrared spectrograph \citep{2019PASP..131e5001W}.  The BOSS and APOGEE spectrographs have a capacity of 500 and 300 fibers respectively, so not every robot will be available for near-infrared targeting.

SDSS-V builds on a heritage of evolution and cooperation between instrument, infrastructure, science, and data teams to obtain and deliver a wide and diverse set of data products to the community.  Operationally, SDSS will see major changes with the inclusion of the FPS instruments, and the teams are currently tackling the many challenges inherent in organizing, optimizing, and deploying a successful SDSS survey operating in a completely new mode.  The work presented here outlines and retires one of the major risks involved with SDSS-V's move to a heavily-overlapping RFP array: designing a strategy for safe and efficient robot trajectory planning during array reconfiguration.

\section{Focal Plane System Layout} \label{sec:layout}

The following subsections describe the layout and kinematics of the SDSS-V RFP array with two benign simplifications: (1) each robot carries a single fiber, and (2) the focal plane is a flat, Euclidean plane.  In deployment, SDSS-V routines will account for both a slight focal plane curvature, and support dual-fiber positioning during trajectory generation.  These SDSS-V hardware-specific details unnecessarily complicate the overall generic formulation of the algorithms we present, so we omit them.

\subsection{Positioner Kinematics} \label{sec:kinematics}

\begin{figure}[!htbp]
\plotone{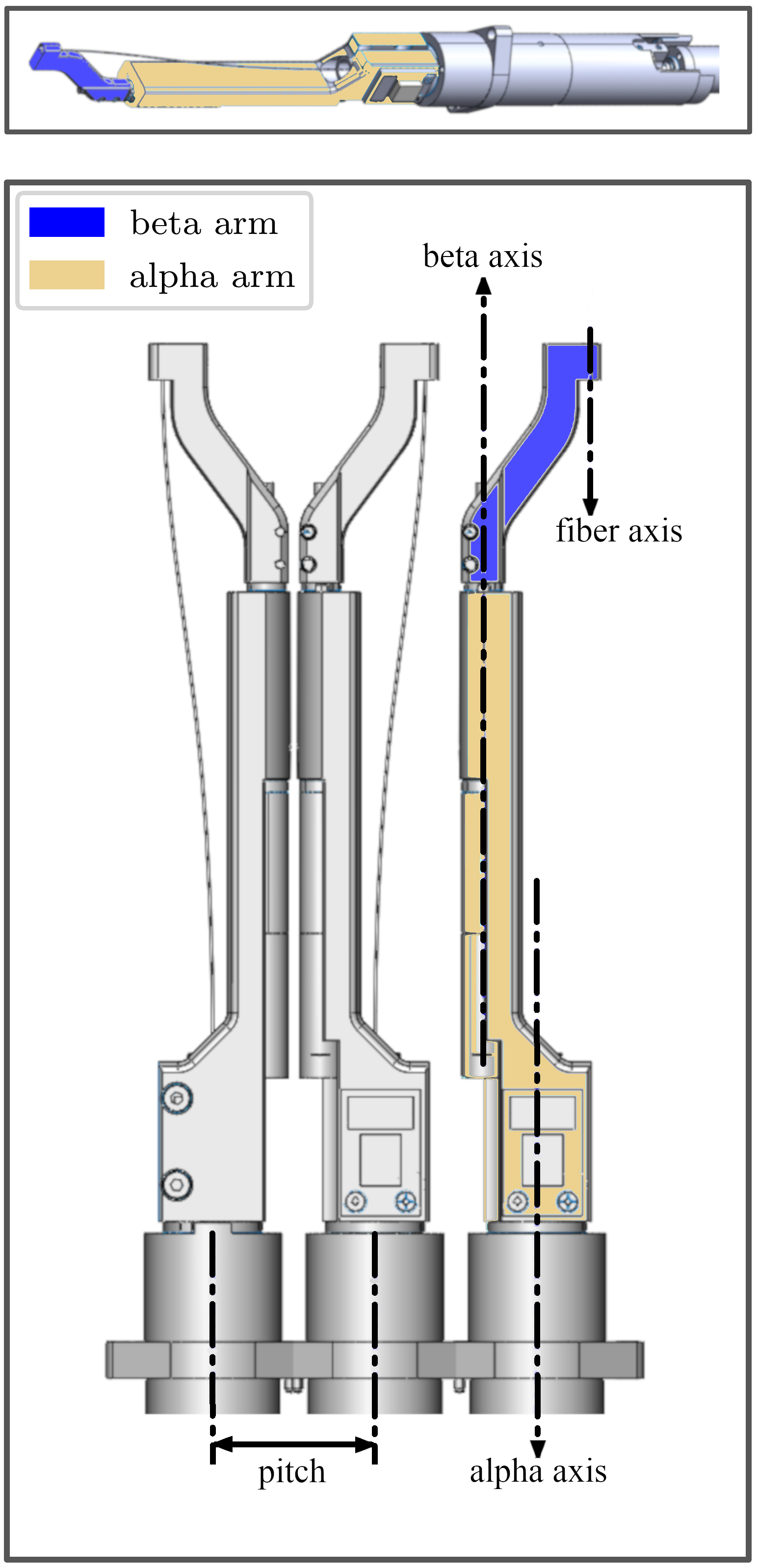}
\caption{Renderings of SDSS-V robots.  The upper panel shows a single RFP unit, the lower panel shows the relative packing of three units at the focal plane viewed edge on at the nominal SDSS-V pitch, where the pitch is the distance between robot centers.  The alpha arm (lower arm) is colored gold, the beta arm (upper arm) is colored blue and holds the optical fiber.  Fiber positioning is accomplished through a rotation of the alpha arm about the alpha axis, and a rotation of the beta arm about the beta axis.  Beta arms risk collision with one another, but alpha arms do not.  The left two positioners in the bottom panel show the level of clearance between alpha arms in an orientation of closest approach.
\label{fig:view1}}
\end{figure}

A rendering of the SDSS-V RFP is shown in the upper inset of Figure \ref{fig:view1}.  The lower panel of this figure shows the relative packing of positioners in the focal plane when viewed edge on. Fiber positioning in the focal plane is achieved through the rotations of two arms about two axes.  The alpha arm (lower arm, indicated in gold) rotates about the alpha axis.  The alpha axis is collinear with the lower body of the robot.  The beta arm (upper arm, indicated in blue), rotates about a beta axis near the edge of the alpha arm.  The distance between the alpha axes of neighboring robots we call the pitch.  The beta arm carries the optical fiber and risks collisions with the beta arms of neighboring robots.  Alpha arms cannot collide with one another.  The left two robots in the lower panel of Figure \ref{fig:view1} are in an orientation of closest approach between alpha arms, showing the tight clearance between them.  Beta arms cannot collide with alpha arms, as they exist in different planes along the optical axis.

\begin{figure}[ht!]
\plotone{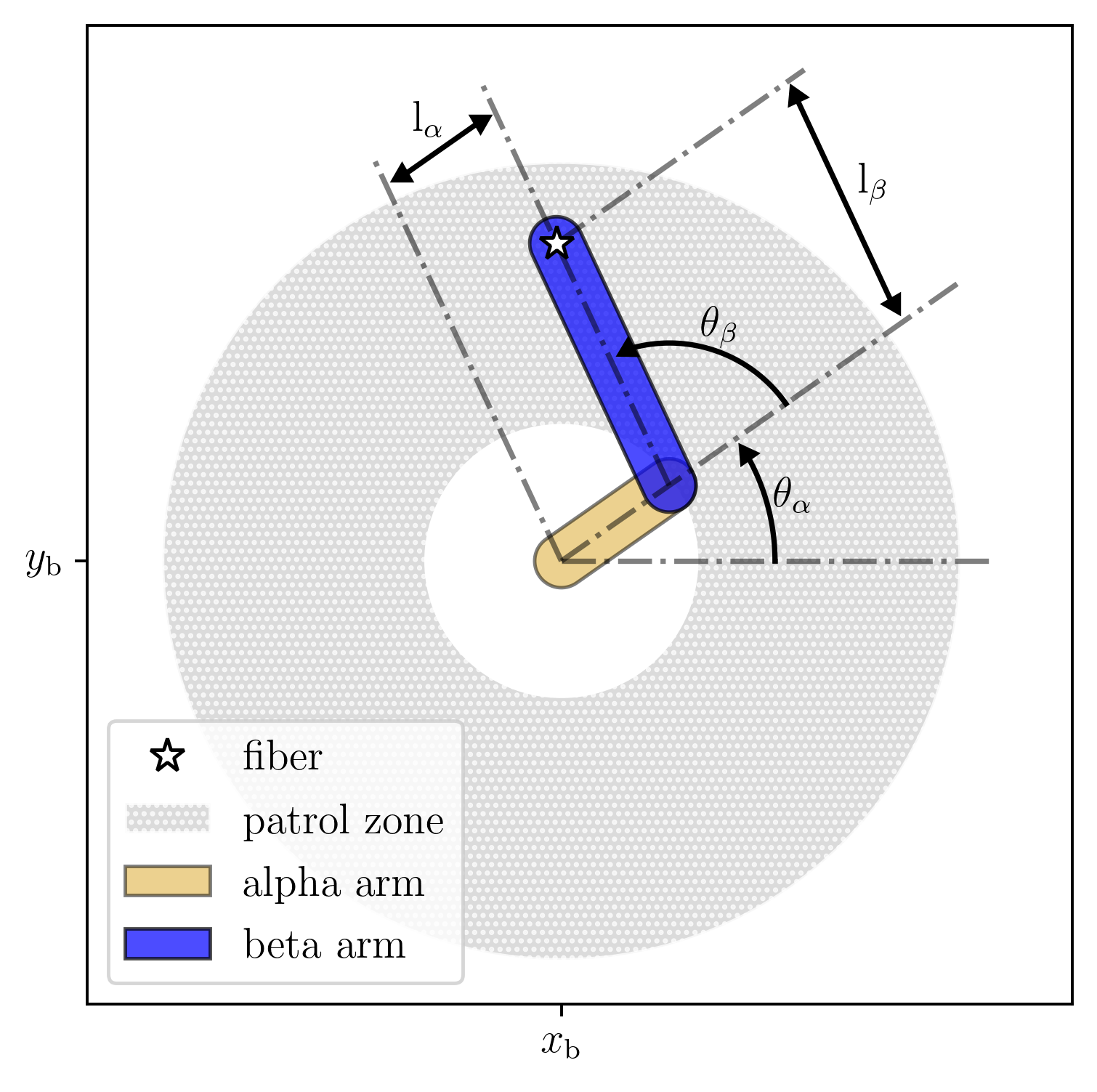}
\caption{A focal plane projection of an SDSS-V robot, showing the annular patrol zone in gray.  Fiber position in the focal plane (white star) is achieved through the specification of two angular rotations $\theta_\alpha$ and $\theta_\beta$ of the alpha and beta arms.  Alpha and beta arms are colored gold and blue.  $\mathrm{l}_\alpha$, $\mathrm{l}_\beta$ indicate the alpha and beta arm lengths.
\label{fig:patrolZone}}
\end{figure}

To visualize how a robot positions a fiber, a focal plane view is helpful.  Figure \ref{fig:patrolZone} shows the focal plane projection of a fiber positioner centered at the robot's xy base position $(x_\mathrm{b}, y_\mathrm{b})$.  Robot arms and rotation axes are indicated on the figure.  The alpha arm length ($\mathrm{l}_\alpha$) is the distance from the alpha axis to beta axis (7.4 mm for SDSS-V).  The beta arm length ($\mathrm{l}_\beta$) is the distance from the beta axis to the fiber center (15 mm for SDSS-V).  The robot may position a fiber anywhere in the annular patrol zone through the specification of the alpha angle ($\theta_{\alpha}$) and beta angle ($\theta_{\beta}$).  This coordinate conversion is:

\begin{equation} \label{eq:1}
\begin{pmatrix}
x_\mathrm{f} \\
y_\mathrm{f}
\end{pmatrix}
=
\begin{pmatrix}
x_\mathrm{b} \\
y_\mathrm{b}
\end{pmatrix}
+
\begin{pmatrix}
\cos \theta_\alpha \quad \cos (\theta_\alpha + \theta_\beta) \\
\sin \theta_\alpha \quad \sin (\theta_\alpha + \theta_\beta)
\end{pmatrix}
\begin{pmatrix}
\mathrm{l}_\alpha \\
\mathrm{l}_\beta
\end{pmatrix}, 
\end{equation}

where $(x_\mathrm{f}, y_\mathrm{f})$ corresponds to the position of the fiber in the xy focal plane.

The physical limits of travel for the SDSS-V positioner are roughly between $-5^\mathrm{o}$ and $365^\mathrm{o}$ for both the alpha and beta axes.  This permits each axis slightly more than one full rotation between hard stops.  Operationally we enforce slightly more restrictive limits of travel during path planning:  each axis must remain in the range $[0^\mathrm{o},360^\mathrm{o})$.

Notice that the orientation in Figure \ref{fig:patrolZone} resembles a human's view of a right arm, and increasing $\theta_\beta$ beyond $180^\mathrm{o}$ would resemble a left arm.  Every point in a positioner's workspace is physically achievable by both a left and right arm conformation.  We will limit targeting to right-armed configurations, though positioners may potentially take on left-armed configurations while maneuvering.  This choice does not impact a positioner's ability to cover the available workspace, but it will eliminate an alternative positioning option during target assignment.  The elimination of the left arm targeting configuration serves two practical purposes in SDSS-V.

The first purpose is to avoid degenerate solutions.  When the FPS system is deployed, a fiber viewing metrology system is used to determine a robot's orientation by measuring the centroid of a back-lit fiber.  If the option of a left arm configuration is eliminated there is only one possible solution for $(\theta_\alpha, \theta_\beta)$ given a measurement of $(x_f,y_f)$.  This is important when considering a malfunctioning robot that cannot report its absolute position.  The robot array will be powered down during science integrations to minimize heat at the focal plane.  If we enforce that robots are only powered down in right arm configurations, this will minimize confusion that could arise due to a malfunction after a power cycle.  While robots remain powered, their absolute positions are frequently reported.

The second purpose is of mechanical nature, rather than mathematical.  We achieve higher fiber positioning accuracy when $0^\mathrm{o} <= \theta_\beta<=180^\mathrm{o}$.  This is due to an interaction between the beta axis preload spring and the direction of torque imparted by the optical fiber itself.  In right arm configurations, the fiber torque and preload work together.  In left arm configurations, the fiber torque and preload oppose each other, leading to slightly higher variance in absolute positioning when operating in especially cold weather.

 \subsection{Positioner Arrangement}

\begin{figure}[ht!]
\plotone{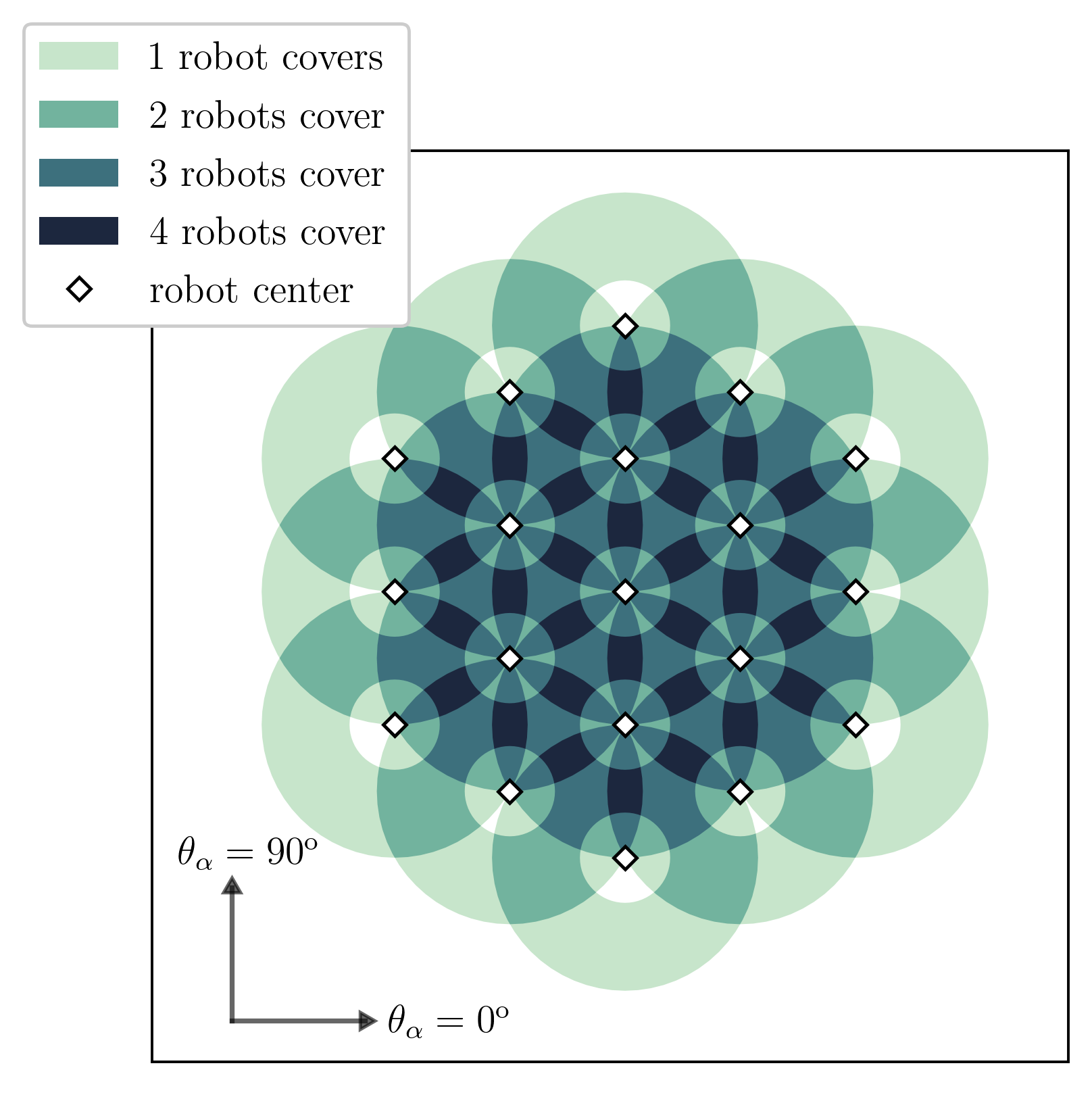}
\caption{A 19 positioner hexagonal array, showing the high packing density of SDSS-V RFPs.  The area is shaded according to the number of positioners that patrol the space.  The six surrounding neighbors may occupy the space at a robot's center (white diamonds).  Generally the area is covered by 3-4 positioners.  Only the array's perimeter is covered by fewer than three robots.  The direction of $\theta_\alpha$ is labeled to indicate the orientation of the robot's coordinate system with respect to the grid.
\label{fig:overlap}}
\end{figure}

To obtain full focal plane coverage the robots are spaced at a pitch of $\mathrm{l}_\alpha + \mathrm{l}_\beta$ in a hexagonal array (22.4 mm for SDSS-V).  This allows a robot's neighbor to patrol its central exclusion zone, leading to heavily overlapping patrol zones between a robot and its neighbors.  Figure \ref{fig:overlap} shows the overlapping patrol zones for an array of 19 SDSS-V positioners, indicating the areas covered by 1, 2, 3, and 4 fibers in the focal plane.  Scaling up the hexagonal array to hundreds of positioners, the majority of the focal surface will be covered by 3 or more fibers.  Only the perimeter of the hexagonal array will have single fiber coverage with some gaps.  The axes in the lower left of the figure indicate the orientation of $\theta_\alpha$ with respect to the hexagonal grid.

The SDSS-V specifications for $\mathrm{l}_\alpha$, $\mathrm{l}_\beta$, and pitch were chosen to maximize the patrol area for each positioner under present constraints of the telescope and spectrograph.  The BOSS spectrograph has a 500 fiber capacity, dictating the total number of robots.  Evenly distributing 500 robots over the telescope's field of view yields a desired pitch of 22.4 mm.  Given the pitch, a 7.4 mm alpha arm length grants the largest possible reach while ensuring that alpha arms cannot physically collide with one another.  With pitch and alpha arm length selected, a 15 mm beta arm length is required to ensure gap-less coverage of the focal plane.  \cite{horler} provide further discussion on the general topic of RFP design and arrangement, including the densely-packed arangement selected for SDSS-V.

\subsection{Collision Formalism} \label{sec:colform}

All physical interference between robots in the grid happens between beta arms.  We use a relatively simple strategy to represent a beta arm geometry in Cartesian space.  The beta arm is modeled by two elements: (1) a line segment and (2) a collision buffer ($\sigma_\mathrm{cb}$).  The beta line segment is constructed between two points $(x_\mathrm{f}, y_\mathrm{f})$, and $(x_\mathrm{e}, y_\mathrm{e})$.  The former point is given in Equation \ref{eq:1}.  The coordinates of the latter point (elbow point) describe the position of the beta axis:

\begin{equation} \label{eq:2}
\begin{pmatrix}
x_\mathrm{e} \\
y_\mathrm{e}
\end{pmatrix}
=
\begin{pmatrix}
x_\mathrm{b} \\
y_\mathrm{b}
\end{pmatrix}
+
\begin{pmatrix}
\mathrm{l}_\alpha\cos \theta_\alpha \\
\mathrm{l}_\alpha\sin \theta_\alpha
\end{pmatrix} 
\end{equation}

The orientation of the beta line segment will depend on a positioner's ($\theta_\alpha, \theta_\beta$) coordinates which vary during motion.  A positioner's base position $(x_\mathrm{b}, y_\mathrm{b})$ and arm lengths $(\mathrm{l}_\alpha$, $\mathrm{l}_\beta)$ remain fixed.

We measure the proximity between two beta arms by computing the minimum distance between beta arm segments.  The procedure for this calculation is provided in Appendix \ref{sec:code} and was adapted from an online source\footnote{\url{http://geomalgorithms.com/a07-_distance.html}}.  We label the minimum distance between beta arm segments for positioners $i$ and $j$ as $\mathrm{D}_{ij}$.

The collision buffer ($\sigma_\mathrm{cb}$) specifies a radius from the beta line segment that safely encloses the physical extent of the beta arm.  We refer to this buffered line segment as the collision envelope, which is a cigar shaped area.  We say robots $i$ and $j$ are collided when:

\begin{equation} \label{eq:3}
\mathrm{D}_{ij} \leq 2\sigma_\mathrm{cb}
\end{equation}

\begin{figure}[ht!]
\plotone{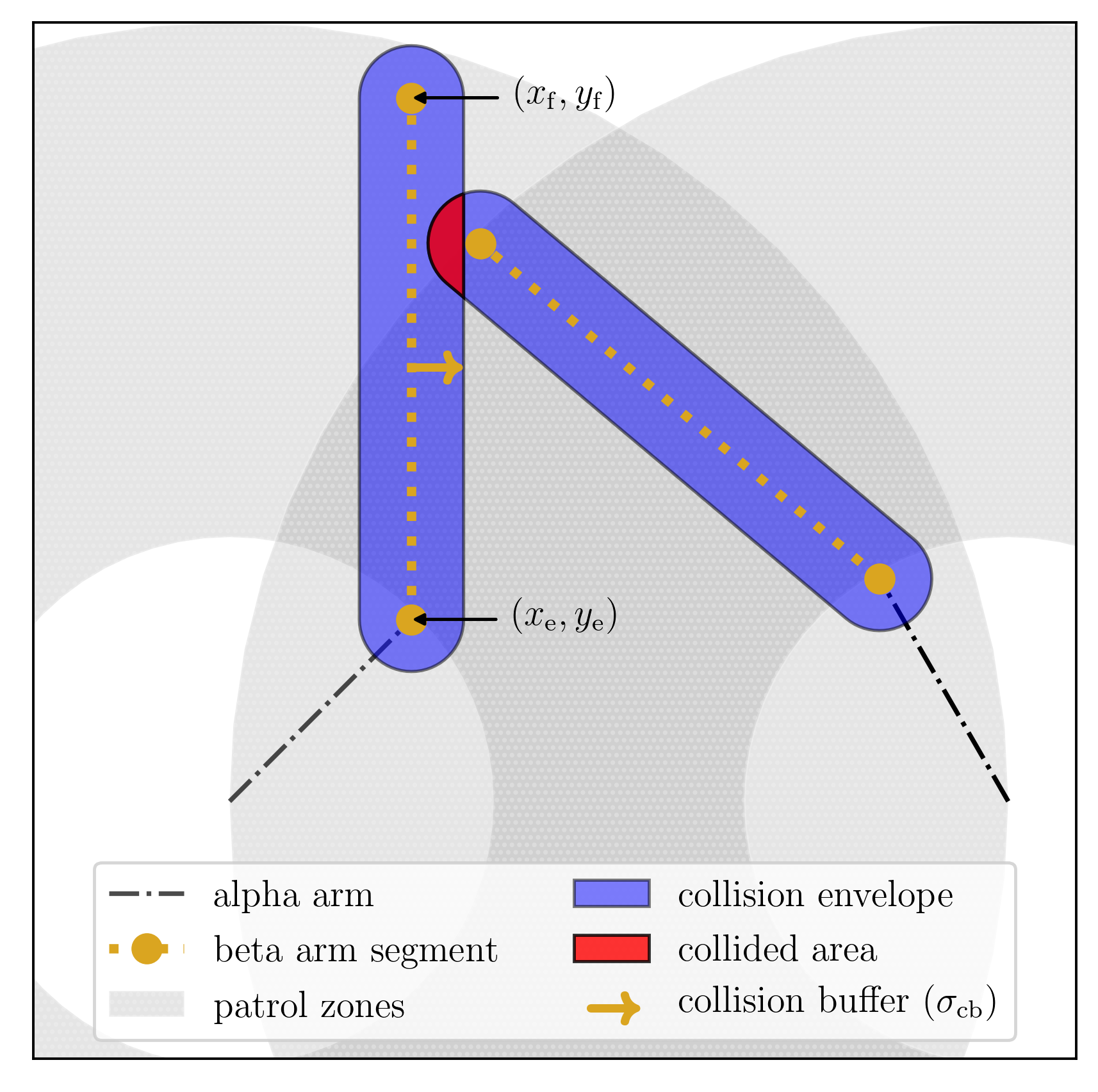}
\caption{Geometric representation of collided beta arms.  The beta arm is described by a line segment constructed of two points $(x_\mathrm{e}, y_\mathrm{e})$, $(x_\mathrm{f}, y_\mathrm{f})$ and a collision buffer $\sigma_\mathrm{cb}$ (gold arrow), where $\sigma_\mathrm{cb} = 1.5$ mm in this view.  The area generated by the line segment and $\sigma_\mathrm{cb}$ is the blue-colored collision envelope, which contains the physical extent of the beta arm.  When collision envelops intersect (Equation \ref{eq:3}, indicated as red in the figure), the two robots are collided.  The overlapping patrol zones are shown in light gray.
\label{fig:colbuffer}}
\end{figure}

Figure \ref{fig:colbuffer} constructs the geometric representation of the beta arm segment (dashed gold line), $\sigma_\mathrm{cb}$ (gold arrow), and collision envelope (blue area) for two colliding RFPs.  The red area in the figure indicates the collided area between the two positioners, where the inequality in Equation \ref{eq:3} is satisfied.

The two dimensional collision envelope we have drawn is conservative when considering the three dimensional shape of the beta arm.  Certain robot orientations would allow the head of a beta arm hover directly above a neighbor's beta axis without suffering a physical collision between beta arms.  Consider the view in Figure \ref{fig:view1} and imagine the central positioner's beta arm floating rightward.  Such an orientation would not be allowed in the focal plane projected view of Figure \ref{fig:colbuffer}, as the area around the elbow joint is encompassed within the collision envelope.  The extent of the collision buffer was chosen to completely allocate the space below a beta arm as belonging to the optical fiber, eliminating any chance of physical interaction between a fiber and a neighboring robot body.  Admittedly, reserving all space below the beta arm is perhaps unnecessarily greedy, as the fiber is tugged closer to the beta arm body when the arm is extended.  The allowance of some amount of safe physical overlap between beta arms could be accomplished by either reducing the length of the beta arm segment to yield space around the elbow, or modeling the beta arm as a segment or curve in three dimensional space.  Permitting beta arm overlap would provide additional real estate for both target assignment and path generation which might prove beneficial.  However, we find the collision buffer as described in this section sufficient for SDSS-V survey operations, and the algorithms we present assume these chosen constraints.

\section{Algorithms} \label{sec:algorithm}
We present two algorithms for the SDSS-V RFP reconfiguration problem: a Greedy Choice (GC) algorithm, and a Markov Chain (MC) algorithm.  Both algorithms carve paths in a stepwise fashion through a series of discrete state transitions by selecting next-state options from a set of small perturbations about the current alpha and beta angles.  The two algorithms vary in their implementation: the GC algorithm selects moves ``greedily" according to a simple heuristic, while the MC algorithm injects a tunable level of randomness into the GC routine.  These algorithms are best classified as Distributed Model Predictive Control (i.e. \citealt{dmpc}), where positioner's choice of move depends only on the current state of its immediate spatial neighbors.  As a heuristic, this simple distributed approach requires minimal computation and results in quickly solved solutions.

The algorithms that follow generically assume a ``perfect" fiber positioner possessing infinite acceleration and no positional uncertainty.  We describe a post-processing strategy that allows for path-adaptation to realistic hardware constraints in Section \ref{sec:deployment}.

\subsection{Definitions} \label{sec:defs}

We first introduce the parameters, metrics, and general ingredients before describing the procedures of the GC and MC algorithms.  We focus our analysis using SDSS-V geometries, although the definitions that follow are generic and can be applied to any layout of RFPs with similar kinematics.

\paragraph{\textbf{Arm Angles}: $\theta_\alpha$, $\theta_\beta$}
The alpha and beta axis angles define the physical orientation of a robot.  The range of motion for each axis is $[0^\mathrm{o}, 360^\mathrm{o})$, as described in Section \ref{sec:kinematics}

\paragraph{\textbf{Arm Lengths}: $\mathrm{l}_\alpha$, $\mathrm{l}_\beta$}
The lengths of the alpha and beta arms.  In this work we use SDSS-V positioner values of $\mathrm{l}_\alpha$ = 7.4 mm and $\mathrm{l}_\beta$ = 15 mm.

\paragraph{\textbf{Collision Buffer}: $\sigma_\mathrm{cb}$}
This parameter sets the relative size of the beta arm, a radial distance from the beta arm line segment (see Section \ref{sec:colform}).  We vary $\sigma_\mathrm{cb}$ between 1.5 and 3.5 mm.  A 1.5 mm $\sigma_\mathrm{cb}$ value exactly encloses the physical envelope of an SDSS-V RFP. A $\sigma_\mathrm{cb}$ between 2 and 2.5 mm maintains a generous amount positioner separation during motion, providing an extra safety buffer for both (1) real-time RFP positional uncertainties along trajectories and (2) path post-processing and simplification.  We touch on these two topics further in Section \ref{sec:deployment}.  In deployment for the SDSS-V FPS, values of $\sigma_\mathrm{cb}$ greater than 2.5 mm are unreasonably large.  However this parameter space will show the potential of these algorithms to achieve high efficiency solutions in regimes more crowded than our specific application.

\begin{figure*}[ht!]
\plotone{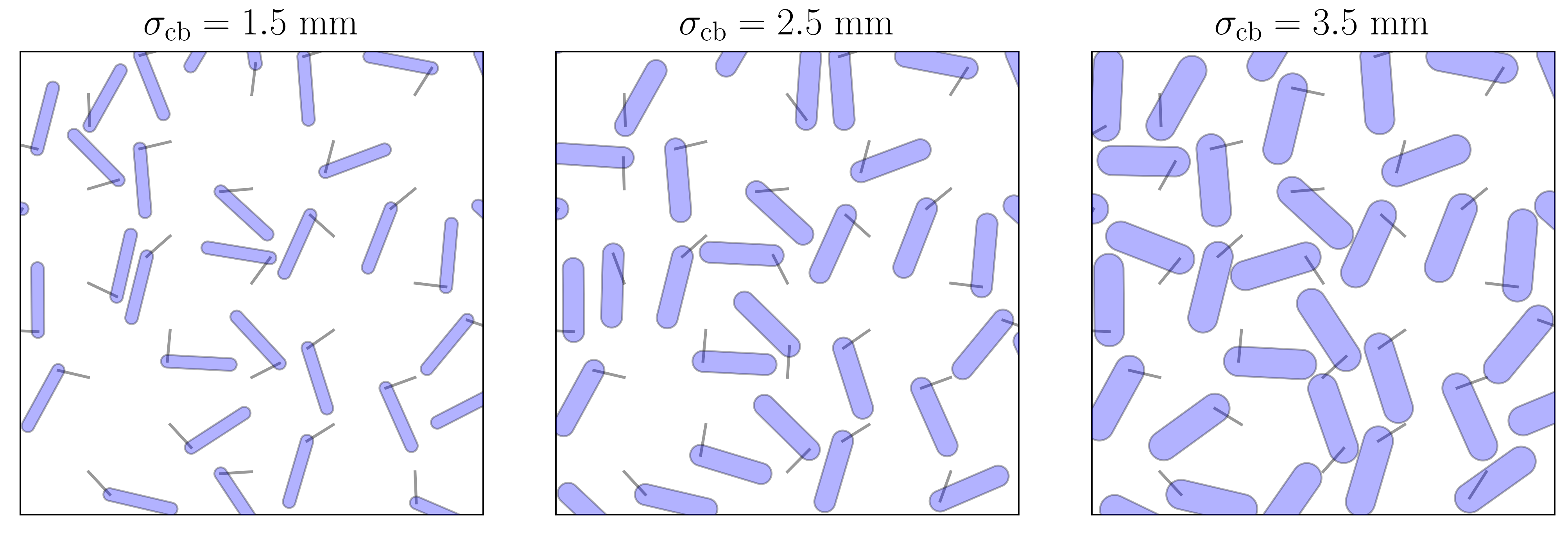}
\caption{Minimum, mean, and maximum $\sigma_\mathrm{cb}$ we consider.  With increasing $\sigma_\mathrm{cb}$ available free space in the focal plane decreases.  $\sigma_\mathrm{cb}$ = 1.5, 2.5, 3.5 mm occupy roughly 6\%, 12\%, and 20\% of the focal plane area for 547 RFP grid.  A $\sigma_\mathrm{cb} = 1.5$ mm represents the physical size of an SDSS-V RFP.  A $\sigma_\mathrm{cb} = 2.5$ mm isolates a large safe zone around an SDSS-V RFP.  A $\sigma_\mathrm{cb} = 3.5$ mm takes up an unreasonable amount of space for an SDSS-V application, but serves to test our algorithms in extremely cramped environments.
\label{fig:cbuffrange}}
\end{figure*}

Figure \ref{fig:cbuffrange} shows the minimum, mean, and maximum $\sigma_\mathrm{cb}$ we investigate, emphasizing the crowding effect with increasing $\sigma_\mathrm{cb}$.  A $\sigma_\mathrm{cb}$ of 1.5, 2.5, and 3.5 mm correspond to roughly 6\%, 12\%, and 20\% of the focal plane area occupied by the collision envelopes of positioners when packed into a SDSS-V layout containing 547 positioners.

\paragraph{\textbf{Robot Centers}: $\vecc{b}_i = (x_{\mathrm{b}}, y_{\mathrm{b}})$}
Base position in the grid for positioner $i$.  For this work we consider hexagonal grids with a pitch of 22.4 mm, matching the pitch of the SDSS-V FPS.  We vary the number of positioners in the grid.

\paragraph{\textbf{Robot Neighbors}: $N_i$}
The set of robot neighbors for positioner $i$.  Two robots are considered neighbors only if they risk collision with one another: when their center-to-center distance is less than twice the sum of their arm lengths and collision buffer $\sigma_\mathrm{cb}$.  If $G$ is the set of all robots in the grid, the set of neighbors for positioner $i$ is:

\begin{equation}\label{eq:neighbors}
N_i = \{j \in G, j \neq i \mid \|\vecc{b}_i - \vecc{b}_j\| \leq 2(\mathrm{l}_\alpha + \mathrm{l}_\beta + \sigma_\mathrm{cb}) \}
\end{equation}

\paragraph{\textbf{Initial Coordinates}: $\vecc{\theta}^\mathrm{\,I}_i = (\theta_{\alpha}, \theta_{\beta})$} The initial alpha and beta angle coordinates for a positioner $i$, a starting point for the routine.  Initial coordinates must be non-colliding.

\paragraph{\textbf{Destination Coordinates}: $\vecc{\theta}^\mathrm{\,D}_i = (\theta_{\alpha}, \theta_{\beta})$} The desired alpha and beta axis angles for a positioner $i$.  The algorithms seek to drive a positioner from its initial coordinates to its destination coordinates, while avoiding collisions with neighbors.  Destination coordinates must be non-colliding.

\paragraph{\textbf{Current Coordinates}: $\vecc{\theta}^\mathrm{\,C}_i = (\theta_{\alpha}, \theta_{\beta})$} The current alpha and beta axis angles for a positioner $i$.  These coordinates evolve with program step, and their history defines the path followed by a positioner.

\paragraph{\textbf{Source Coordinates}: $\vecc{\theta}^\mathrm{\,S}_i = (\theta_{\alpha}, \theta_{\beta})$} The source coordinates specify the alpha and beta axis angles for a positioner to receive light from an astronomical source.  In this work a robot's source coordinates are drawn uniformly from the annular fiber patrol zone.  Source coordinates must be non-colliding.  All source coordinates are right-armed (Section \ref{sec:kinematics}).  If an initial source coordinate assignment creates a collision with a previously assigned robot, source coordinates are redrawn iteratively until the collision vanishes.  In a \textbf{forward} path solution, the source coordinates for positioner $i$ are the destination coordinates, $\vecc{\theta}^\mathrm{\,D}_i$.  In a \textbf{reverse} path solution, the source coordinates serve as the initial coordinates, $\vecc{\theta}^\mathrm{\,I}_i$.

\paragraph{\textbf{Angular Step}: $\Delta\theta$}
The angular step parameter specifies the step size of the routine: a maximum angular perturbation of a robot's current coordinates $\vecc{\theta}^\mathrm{\,C}$ at each program step.  This parameter may be converted into a time step, $\Delta t$ by:

\begin{equation}\label{eq:ts}
\Delta t = \frac{\Delta \theta}{\dot{\theta}},
\end{equation}

where $\dot{\theta}$ is the angular speed of the positioner's alpha and beta axes (30 deg/sec for SDSS-V RFPs).  Smaller angular step values produce more densely sampled paths at the cost of increased program runtime.

\paragraph{\textbf{Maximum Steps}} The maximum number of steps the routine will run for.  For this work we set a limit of $1000^{\rm{o}}/\Delta\theta$ steps.  This corresponds to 1000 degrees of motion, or roughly 33 seconds of motion assuming a 30 degree/sec angular speed for an SDSS-V RFP.

\paragraph{\textbf{Minimum Approach Distance}}
This distance designates the the closest allowed approach between any two beta arm segments in the routine.  For a positioner $i$ with a set of neighbors $N_i$ we enforce the inequality:

\begin{equation} \label{eq:4}
\mathrm{min}\{j\in N_i \mid D_{ij} \} > 2\sigma_\mathrm{cb} + \mathrm{MD},
\end{equation}

where $\mathrm{MD}$ is a bound on the maximum displacement for a beta arm in a single program step.  We choose a conservative value for $\mathrm{MD}$: the displacement of the fiber $(x_\mathrm{f}, y_\mathrm{f})$ with beta arm at full extension ($\theta_\beta = 0$), when rotated 2$\Delta\theta$ about the positioner's central axis:

\begin{equation} \label{eq:md}
\mathrm{MD} = (\mathrm{l}_\alpha + \mathrm{l}_\beta)\sin(2\Delta\theta)
\end{equation}

The inclusion of the maximum displacement term protects against ``tunneling collisions", where one positioner may completely jump through a collided orientation in a single step.  Smaller values of $\Delta\theta$ permit closer approaches between positioners, and thus allow more options for maneuvering.

\paragraph{\textbf{Cost}: $C_i$}
The cost metric is minimized throughout the routine for each positioner.  It measures a distance between a positioner's destination coordinates ($\vecc{\theta}^\mathrm{\,D}_i$) and current coordinates ($\vecc{\theta}^\mathrm{\,C}_i$).  Note this metric is not equal to the Euclidean distance between a fiber and its destination in the xy focal plane.  For a positioner $i$ we define the cost as:

\begin{equation}\label{eq:5}
    C_i = \| \vecc{\theta}^\mathrm{\,C}_i - \vecc{\theta}^\mathrm{\,D}_i \|
\end{equation}

When the cost is zero, the positioner has arrived at its destination.

\paragraph{\textbf{Energy}: $E_i$}
Energy is a measure of local crowding: the sum of inverse square distances to a set of neighbors.  We define the energy for a positioner $i$ with neighbors $N_i$ as:

\begin{equation}\label{eq:6}
    E_i = \sum_{j \in N_i} \left( \frac{1}{D_{ij}} \right)^2
\end{equation}

This metric is only used in the MC algorithm.

\paragraph{\textbf{Phobia}}
Phobia is a scalar between 0 and 1, used only in the MC algorithm.  It represents the weighting of relative importance between the energy metric $E_i$ and the cost metric $C_i$.  A weight of 0 produces a positioner insensitive to crowding.  A phobia weight of 1 produces a positioner only sensitive to crowding and it will seek to separate from neighbors above all else.

\paragraph{\textbf{Greed}}
Greed is a scalar between 0 and 1 used only in the MC algorithm.  It represents how much weight a positioner will place on taking the best move when minimizing the $E_i$ or $C_i$ metrics.  A greed value of 1 introduces no stochasticity in a positioner's move selection sequence, while a greed value of 0.5 resembles a random, undirected walk.  A greed value of 0 produces a robot that will remain stationary.

\paragraph{\textbf{Neighbor Encroachment}}
Neighbor encroachment is a proximity threshold that is only considered in the MC algorithm.  A positioner $i$ with neighbors $N_i$ will remain stationary only while its cost $C_i$ = 0 and the following inequality is satisfied:

\begin{equation} \label{eq:7}
    \mathrm{min}\{j\in N_i \mid D_{ij} \} > 2\sigma_\mathrm{cb} + 3\mathrm{MD}
\end{equation}

This provides a similar constraint to the minimum approach distance outlined in Equation \ref{eq:4}, but is sensitive over a slightly longer distance.  This will cause any robot currently at rest at its destination to be kicked awake by any neighboring robot entering its horizon.

\subsection{The Greedy Choice Algorithm}
The GC algorithm  is a stepwise process.  At each iteration, each robot $i$ in a grid of $G$ is visited in turn and presented with a set of nine options from which to select its next state.  The options consist of the combination of \{$-\Delta\theta, 0, \Delta\theta$ \} perturbations applied to the current alpha and beta coordinates $\vecc{\theta}^\mathrm{\,C}_i$.  Perturbations are modified if limits of travel are violated or destination coordinates are overshot.  The robot will choose the option that both (1) minimizes its cost (Equation \ref{eq:5}), and (2) satisfies the minimum approach criterion (Equation \ref{eq:4}).  The routine terminates when either (1) all robots have reached their destination coordinates ($\sum_{i\in G}C_i = 0$) or (2) the max iteration limit is reached.

\subsection{The Markov Chain Algorithm}
The MC algorithm is an extension of the GC algorithm.  In contrast to the GC algorithm, the MC algorithm selects subsequent states probabilistically rather than greedily, where the level of stochasticity is controlled by the greed parameter.  Additionally, we construct a control policy that jointly minimizes both the cost and energy metrics, where relative weighting between cost and energy is controlled by the phobia parameter.  Greed, phobia, cost, and energy are described in Section \ref{sec:defs}.

At each program step a positioner must first select whether or not to consider a move.  If the positioner is at its destination coordinates and the inequality in Equation \ref{eq:7} is satisfied, it will remain stationary until the next program step, otherwise it will consider a move.

If a positioner considers a move, it must next choose a metric to minimize: either cost or energy.  The probability of selecting the energy metric is equal to the phobia parameter.  With a metric selected, a positioner will visit each of the nine next-state candidates in a random order, where these options are the combinations of \{$-\Delta\theta, 0, \Delta\theta$ \} perturbations applied to the current alpha and beta coordinates. Perturbations are modified if limits of travel are violated or destination coordinates are overshot.  As each state is visited, a positioner will choose to accept or reject the proposed state.  A positioner accepts the proposed state with a probability equal to the greed parameter if the following two criteria are met: (1) this state represents the minimum value of the selected metric seen at this step (by measure of Equation \ref{eq:5} or \ref{eq:6}), and (2) this state satisfies the minimum approach criterion (Equation \ref{eq:4}). If no state is selected, it defaults to remain in its current state until the next program step.

The routine terminates when either (1) all robots have reached their destination coordinates ($\sum_{i\in G}C_i = 0$) or (2) the max iteration limit is reached.

The GC algorithm represents perhaps the simplest possible control law in the framework we have described.  The MC algorithm extends upon the GC algorithm through the inclusion of two additional features: (1) the injection of stochasticity in a robot's decision making process, and (2) a mechanism for robots to sense and avoid crowded spaces.  The greed and phobia parameters provide a tuning knob for these MC features.  When greed approaches 1 and phobia approaches 0, the MC algorithm becomes identical to the GC algorithm.  In the analysis that follows, we fix greed and phobia parameters to 0.9 and 0.3.  This choice of MC parameter setting is not necessarily intended to represent an optimal tuning, but rather to provide a point of comparison between the two approaches.

\vspace{\baselineskip}
\noindent A complete pseudocode implementation for the GC/MC routine is provided in Appendix \ref{sec:code}.

\section{Analysis} \label{sec:analysis}
We begin our analysis using a grid of 19 positioners with $\sigma_\mathrm{cb}$ = 3.5 mm.  This small, crowded grid will serve to both visualize the algorithm's behavior, and motivate a reverse path solve strategy.  Further analysis will implement the reverse solve strategy in large grids, comparing behavior between the GC and MC algorithms at various parameter settings and grid sizes.  We largely focus on 547 positioner grids, as this matches the size of SDSS-V FPS layout\footnote{As described in Section \ref{sec:sdss-v}, the FPS will carry only 500 robots.  For simulations here we elect to fill every available spot with a positioner.}.

SDSS-V's Kaiju\footnote{\url{https://github.com/sdss/kaiju}} package is an open source Python-wrapped C++ package implementing the collision avoidance algorithms.  The results we present here were obtained using the Kaiju package (tag 0.5.0) compiled with clang++ (version 10.0.1) using an -O3 optimization flag running in a single thread on a 2.9 GHz Intel Core i9 CPU.

\subsection{Motivating the Reverse Solver}\label{sec:forrev}

The performance of the GC and MC algorithms can be highly dependent on direction of solved motion.  Consider two array configuration states between which we want find collision-less paths for all robots.  The first state $\vecc{\theta}^\mathrm{\,F}_i = (0^{\rm{o}}, 180^{\rm{o}})$ represents a ``folded" state in which all robot arms are retracted and aligned in a lattice-like orientation.  The second state is a randomized source coordinate configuration $\vecc{\theta}^\mathrm{\,S}_i$ which approximates an orientation of RFPs positioned to receive light from astronomical sources in a field.  When solving the \textbf{forward} path from initial coordinates $\vecc{\theta}^\mathrm{\,F}$ to destination coordinates $\vecc{\theta}^\mathrm{\,S}$ a vast number of positioners will deadlock and never their destination.  However, if we swap the initial and destination coordinates and solve the \textbf{reverse} path $\vecc{\theta}^\mathrm{\,S} \rightarrow \vecc{\theta}^\mathrm{\,F}$, we find our routines yield near-perfect convergence of robots to their desired destinations.

To demonstrate the forward and reverse solutions we use: the GC algorithm, a 19 positioner grid, and a step size $\Delta_\theta = 0.5$ deg.  We set $\sigma_\mathrm{cb} = 3.5$ mm to simulate an abnormally large level interference between positioners.  Non-colliding source coordinates for each positioner are selected randomly from their respective patrol zones.  Initial coordinates are set such that the robots begin their journey from a folded position, $\vecc{\theta}^\mathrm{\,I}_i = \vecc{\theta}^\mathrm{\,F}_i$.  Destination coordinates are set to be the source coordinates $\vecc{\theta}^\mathrm{\,D}_i = \vecc{\theta}^\mathrm{\,S}_i$.

\begin{figure*}[ht!]
\plotone{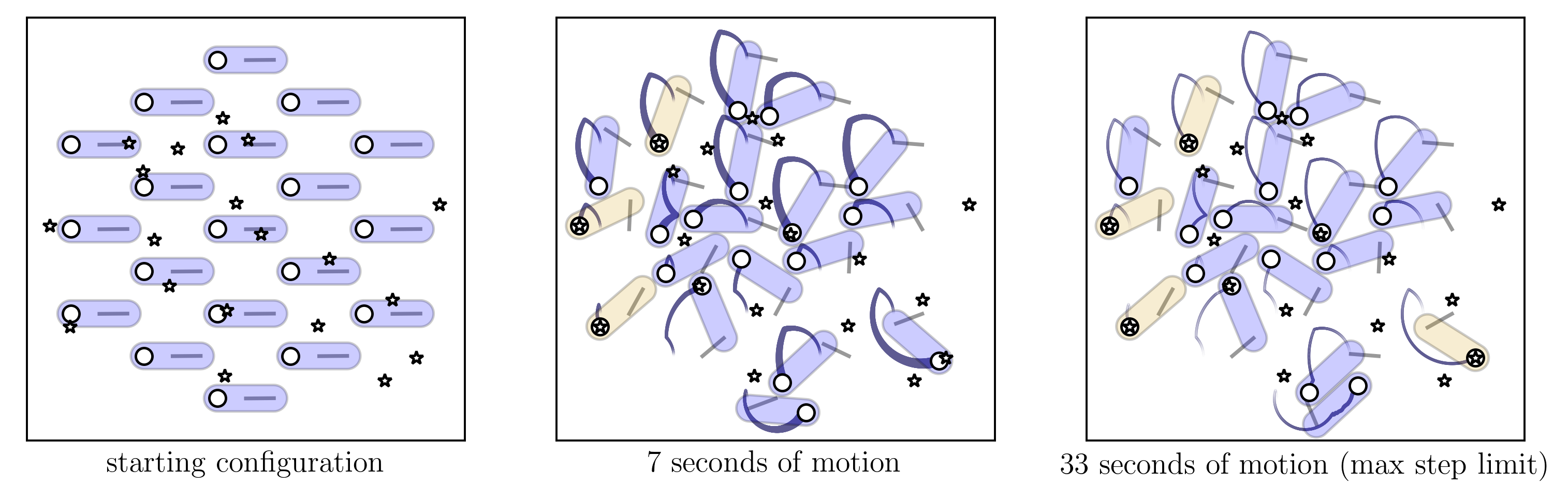}
\caption{A 19 positioner grid solved with the GC algorithm in the \textbf{forward} direction using $\Delta_\theta = 0.5$ deg.  We set $\sigma_\mathrm{cb} = 3.5$ mm to simulate an environment more crowded than SDSS-V, which increases the optimization challenge.  Panels from left to right show the starting configuration, an intermediate state of motion, and the final state.  Stars indicate source coordinates and open circles indicate the fiber.  When a star aligns with its fiber, the robot has reached its destination, and is colored tan.  Streaks behind the fibers indicate the path followed by a fiber through previous program steps.  In this example, the maximum step limit is hit with only 4 positioners reaching their destination (right panel), the remainder of positioners are in deadlock.
\label{fig:forward}}
\end{figure*}

Figure \ref{fig:forward} shows three snapshots in robot-travel time when forward motion is propagated according to the GC algorithm.  The first panel shows the initial folded state of the positioners, the second panel shows an intermediate state in the routine, and the third panel shows the final state.  Stars indicate the source coordinate locations, circles indicate the fiber position at the end of the beta arm.  When the fiber is aligned with the source, the positioner has reached its destination.  Robots at their destination are colored tan, robots in motion are colored blue.  Curved streaks drawn behind the fiber indicate the path the fiber has followed through previous program steps.

In this example, only 4 of 19 positioners arrive at their destination before the routine's iteration limit is reached.  The remaining positioners are in a deadlocked state where all progress is halted due to inability of robots to navigate beyond their collective blockade.

The next demonstration uses identical configurations to those of the previous example with the following modification: initial coordinates are set to the source coordinates $\vecc{\theta}^\mathrm{\,I}_i = \vecc{\theta}^\mathrm{\,S}_i$, and destination coordinates are set to the folded configuration $\vecc{\theta}^\mathrm{\,D}_i = \vecc{\theta}^\mathrm{\,F}_i$.  When solved with the GC algorithm, robots will begin from a source coordinate orientation, and carve out paths toward a folded configuration.  In this case we are solving exactly the same problem as the previous example (a path between a folded and source orientation), by simply reversing the direction of robot-arm propagation.

\begin{figure*}[ht!]
\plotone{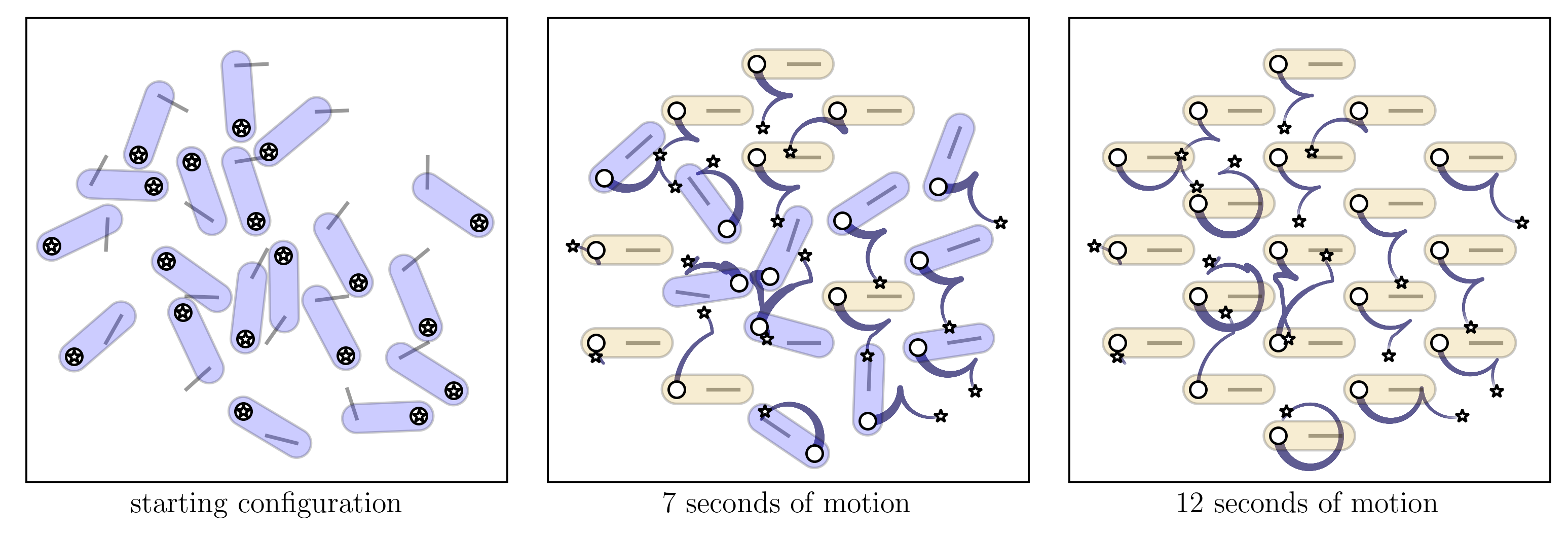}
\caption{A \textbf{reverse} direction GC solution to the configuration presented in Figure \ref{fig:forward}.  The initial state of the routine is set to the source coordinate configuration (left panel) in which all fibers are aligned with astronomical targets.  The program steps toward a folded destination (right panel).  Here all 19 positioners successfully navigate to the folded destination in 12 seconds of motion.  A forward path may be obtained by reflecting the reversely solved path in time, providing a route for positioners to go out and back from a folded state when visiting this field of science targets.
\label{fig:reverse}}
\end{figure*}

Figure \ref{fig:reverse} shows the results of the reverse GC solve, an analogous sequence to that presented in Figure \ref{fig:forward}.  Positioners begin aligned with source coordinates, and step toward a folded destination.  In this experiment we find that all positioners successfully navigate to their destinations after 12 seconds of motion.  This result is striking when contrasted with the forward example in which only a small subset of positioners were able to obtain their desired destination.  This pair of examples demonstrates that the reverse path strategy is a surprisingly good approach.

The irreversible behavior of the algorithm raises eyebrows (the authors' included), yet it provides unique leverage in solving our path routing problem.  Consider the two following points.  First, paths are trivially reversible.  If a reverse path solution is found, we can obtain a forward path solution by simply reflecting trajectories of all robots.  Second, by building all paths between any source coordinate state and a common folded state, transitioning between any two source coordinate states is also trivial given that reconfigurations always route through the folded state.

The reverse path approach was discovered somewhat serendipitously through experimentation, and motivated by the following thought experiment: is it easier to tie a complex knot, or untie a complex knot?  Presumably the latter task is easier, and if one carefully recorded the steps taken while performing the latter task, then the reversal of those steps yeilds a solution to the former task.  Regrettably, we do not have a theoretical justification as to why the reverse process works so well in the context of our problem, though we find the behavior interesting and provide some further discussion in Section \ref{sec:discussion}.  In further sections we rely on statistics gathered from large numbers of simulations to gain confidence in the quality of our solutions.

\subsection{Measuring Efficiency with Source Coordinate Replacement}\label{sec:replace}

\begin{figure}[ht!]
\plotone{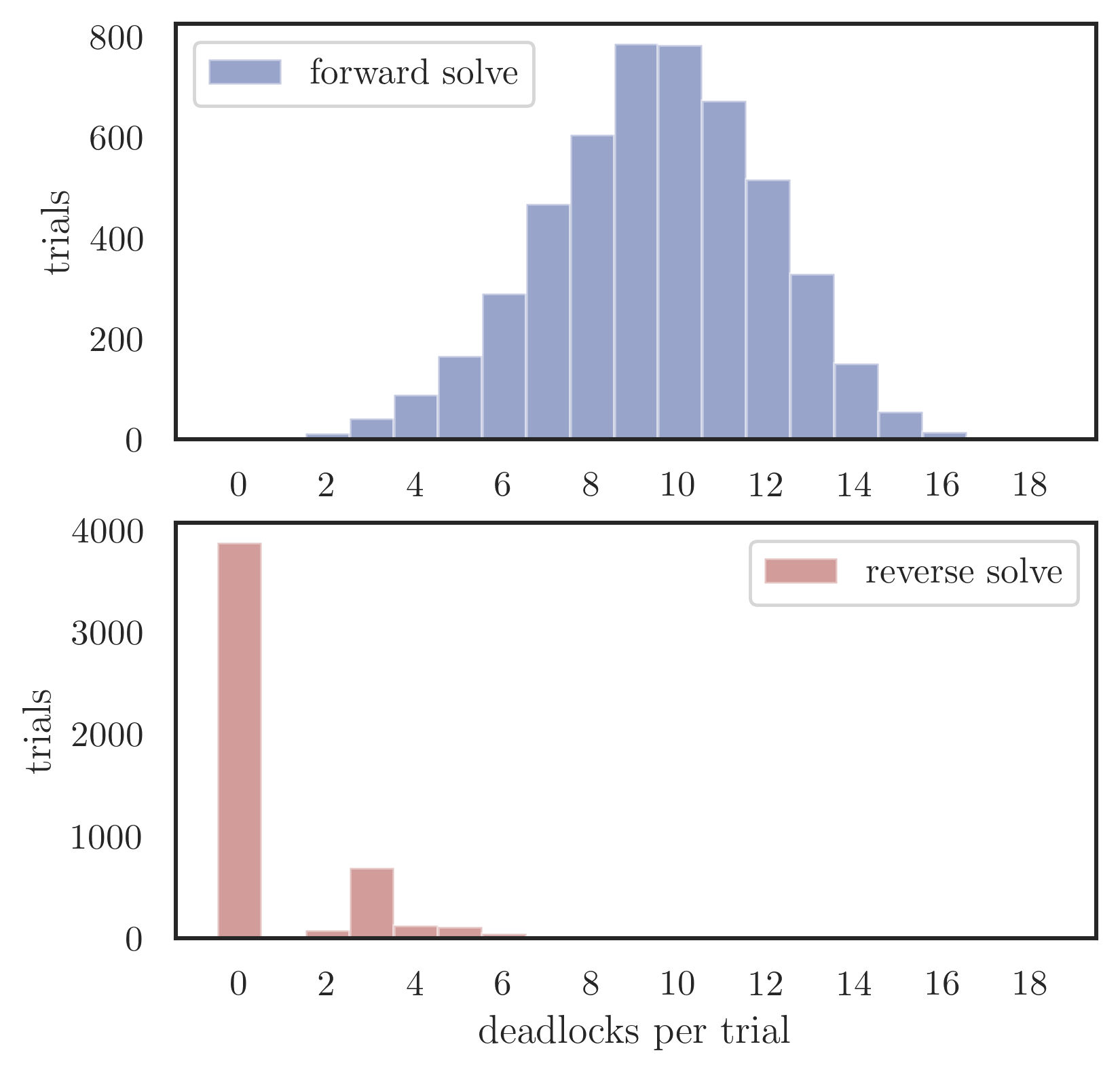}
\caption{Histograms comparing deadlock frequencies between the forward and reverse solve methods for abnormally crowded grids.  Histograms are generated from 5000 trials using the GC algorithm, a 19 positioner grid, $\sigma_\mathrm{cb} = 3.5$ mm, and $\Delta_\theta = 0.5$ deg.  Essentially we have repeated the trajectory comparisons between Figures \ref{fig:forward} and \ref{fig:reverse} with source coordinates randomized between trials.  In forward solutions (upper panel), every trial experienced a deadlock, and the typical deadlock involved many positioners.  For reverse solutions (lower panel), the large majority of trials suffered no deadlock, and those that did typically involved only a few positioners.  From this example, the reverse path solution is well motivated.
\label{fig:revForBig}}
\end{figure}

Here we formulate a statistical measure of algorithmic efficiency for the layout described in Section \ref{sec:forrev}.  We repeat GC forward/backward pair analysis in a set of 5000 trials, where source coordinates are randomized between trials.  Figure \ref{fig:revForBig} shows the resulting histogram of positioner deadlock frequencies for both strategies.  The forward strategy will always result in deadlock that typically involves half of the positioners, whereas the reverse strategy only rarely suffers a deadlock.  In these rare cases of reverse path deadlock, only a few positioners are involved.  The chance of deadlock in the reverse solution is small, but non-negligible.

For the reverse solution to be viable, we require a deadlock-free path for every positioner in the grid.  The full reconfiguration sequence from source coordinates A to subsequent source coordinates B requires two steps: (step 1) the array moves from A to an intermediate folded state, then (step 2) the array moves from the folded state to B.  If any single positioner deadlocks during reverse path generation, the array has not made it to the transition state, and thus the set of paths cannot be used.  Having a strategy for eliminating deadlocks is a necessary feature for the reverse path solver.

We adopt a brute-force method of deadlock resolution.  If a grid is deadlocked, we randomly select a single deadlocked robot, replace its source coordinates, and re-run the path generator.  Replacing a source coordinate results in the loss of the original astronomical source.  In large grids of positioners, deadlocks are typically isolated in small groups throughout the array.  For these cases we may simultaneously replace one robot from each deadlocked group, which minimizes the number of iterations required of the path generator, and improves the total runtime.

\begin{figure*}[ht!]
\plotone{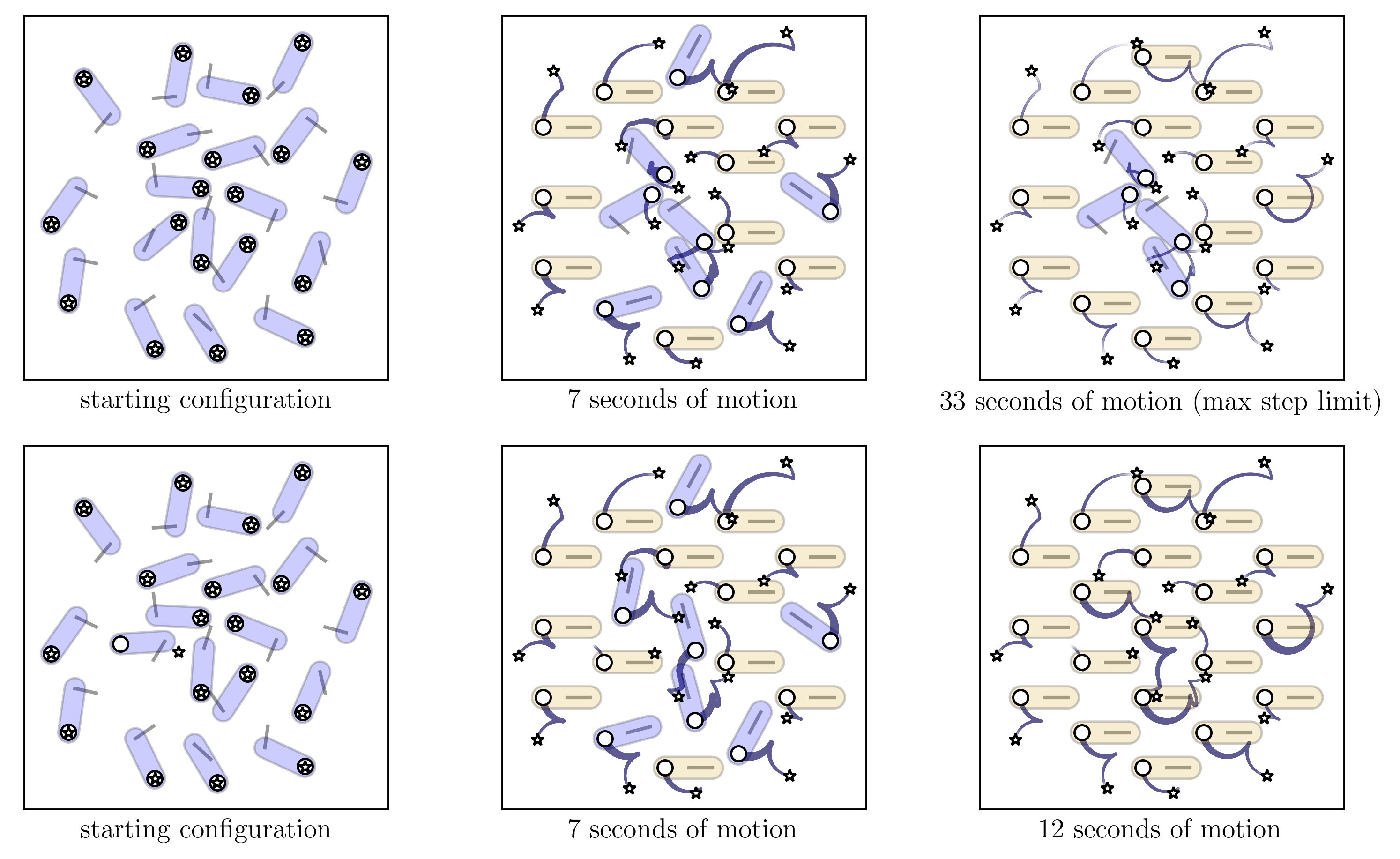}
\caption{We illustrate a brute force method for deadlock resolution using the GC algorithm, a 19 positioner grid, $\sigma_\mathrm{cb} = 3.5$ mm, and $\Delta_\theta = 0.5$ deg.  The top three panels show an example sequence of reversely solved motion that results in a 4 positioner deadlock.  The bottom three panels show an analogous sequence after a single positioner's initial coordinates have been replaced.  The bottom sequence converges in 12 seconds of motion at the loss of a single astronomical target.  Arriving at a fully converging grid is necessary to deploy a reverse path generator, and thus our procedure requires an iterative search for viable initial conditions when a deadlock is present.  When deadlocks are infrequent, this introduces small computational overhead.
\label{fig:replace}}
\end{figure*}

This brute force replacement strategy is illustrated in Figure \ref{fig:replace}.  The top three panels show a sequence of reverse-solved motion that results in a four positioner deadlock.  The lower three panels show the results obtained after replacing a single positioner's source coordinates to a new random position.  The array now completely converges to a folded orientation in 12 seconds of motion, thus resolving four wedged robots with a single replacement.

We define an efficiency for the reverse GC and MC algorithms in terms of the required number of replacements to solve a grid.  For a grid with total number of positioners $n_{G}$ requiring source coordinate replacements for a subset of positioners $n_{R}$, the efficiency is:

\begin{equation}
\textrm{efficiency} = \frac{n_{G} - n_{R}}{n_{G}}
\end{equation}

Note that we count $n_{R}$ as number of robots requiring a source coordinate replacement.  If a single robot is iteratively tried with multiple sets of source coordinates, we count this as a single replacement.  This measure of efficiency is the ratio of astronomical targets obtained to astronomical targets assigned. For the example shown in Figure \ref{fig:replace}, the efficiency is $(19-1)/19 \sim 0.95$.

In our trials we generally (but not necessarily) find that the efficiency is higher than the initial convergence of the grid, where we call convergence the ratio of positioners achieving the destination to the total number of positioners in the grid.  In the case shown in Figure \ref{fig:replace}, the initial convergence (top panel sequence) is $(19-4)/19 \sim 0.79$.  The efficiency increase with respect to convergence is due to the fact that a single replacement will often resolve more than one deadlock.  We use efficiency (rather than convergence) to measure algorithmic performance as it is a more relevant statistic in the context of astronomical surveys.

\subsection{Efficiency vs. Crowding} \label{sec:eff}

In a cocktail party setting, navigation becomes more challenging as the free space in the room shrinks.  The $\sigma_\mathrm{cb}$ parameter is a proxy for either beta arm relative size or collision safety distance, and as $\sigma_\mathrm{cb}$ increases, available space for motion in the grid decreases.  Here we investigate the path routing performance of the reverse-solve GC and MC algorithms for a range of $\sigma_\mathrm{cb}$ values between 1.5 and 3.5 mm to simulate varied levels of crowding.

We use a grid of 547 positioners and set a step size $\Delta_\theta = 0.1$ deg.  This step size limits robot arm perturbations (Equation \ref{eq:md}) to less than 100 microns at each iteration.  We run 2000 trials at each $\sigma_\mathrm{cb}$ for each algorithm, amounting to 36000 trials in total.  Initial coordinate assignments are randomized in all trials.  We assign folded destination coordinates $\vecc{\theta}^\mathrm{\,D} = (10^\mathrm{o}, 170^\mathrm{o})$ for all positioners.  These destination coordinates leave the positioners with $\pm 10^{\mathrm{o}}$ of free travel in both alpha and beta axes when they arrive at their destination.  This choice of parking spot benefits the MC algorithm, in which positioners at rest may be kicked awake when a neighbor encroaches (Equation \ref{eq:7}), allowing for a larger set of evasive options in these situations.  The parameter settings for this experiment are summarized in Table \ref{tab:sim1}.

\begin{table}[h]
\caption{Simulation 1 - varied crowding} \label{tab:sim1}
\begin{tabular}{rl}
\hline\hline 
Parameter             & Values   \\
\hline
n robots & 547        \\
n trials \footnote{trials split evenly amongst the varied $\sigma_\mathrm{cb}$ settings and algorithm type}               & 36000      \\
$\mathrm{l}_\alpha$ (mm)                   & 7.4   \\
$\mathrm{l}_\beta$ (mm)                   & 15   \\
$\vecc{\theta}^\mathrm{\,I}$ (deg)                   & random, right-armed   \\
$\vecc{\theta}^\mathrm{\,D}$ (deg)                   & (10,170) \\
pitch (mm)             & $\mathrm{l}_\alpha + \mathrm{l}_\beta = 22.4$    \\
$\Delta_\theta$ (deg)                   & 0.1        \\
max steps               & $1000^\mathrm{o}/\Delta_\theta$        \\
$\sigma_\mathrm{cb}$ (mm)                   & $\{1.5, 1.75, 2, 2.25, 2.5, 2.75, 3, 3.25, 3.5\}$        \\
greed                 & 0.9        \\
phobia                & 0.3 \\
algorithm & $\{\mathrm{GC}, \mathrm{MC} \}$ \\
\hline
\end{tabular}
\end{table}

\begin{figure}[ht!]
\plotone{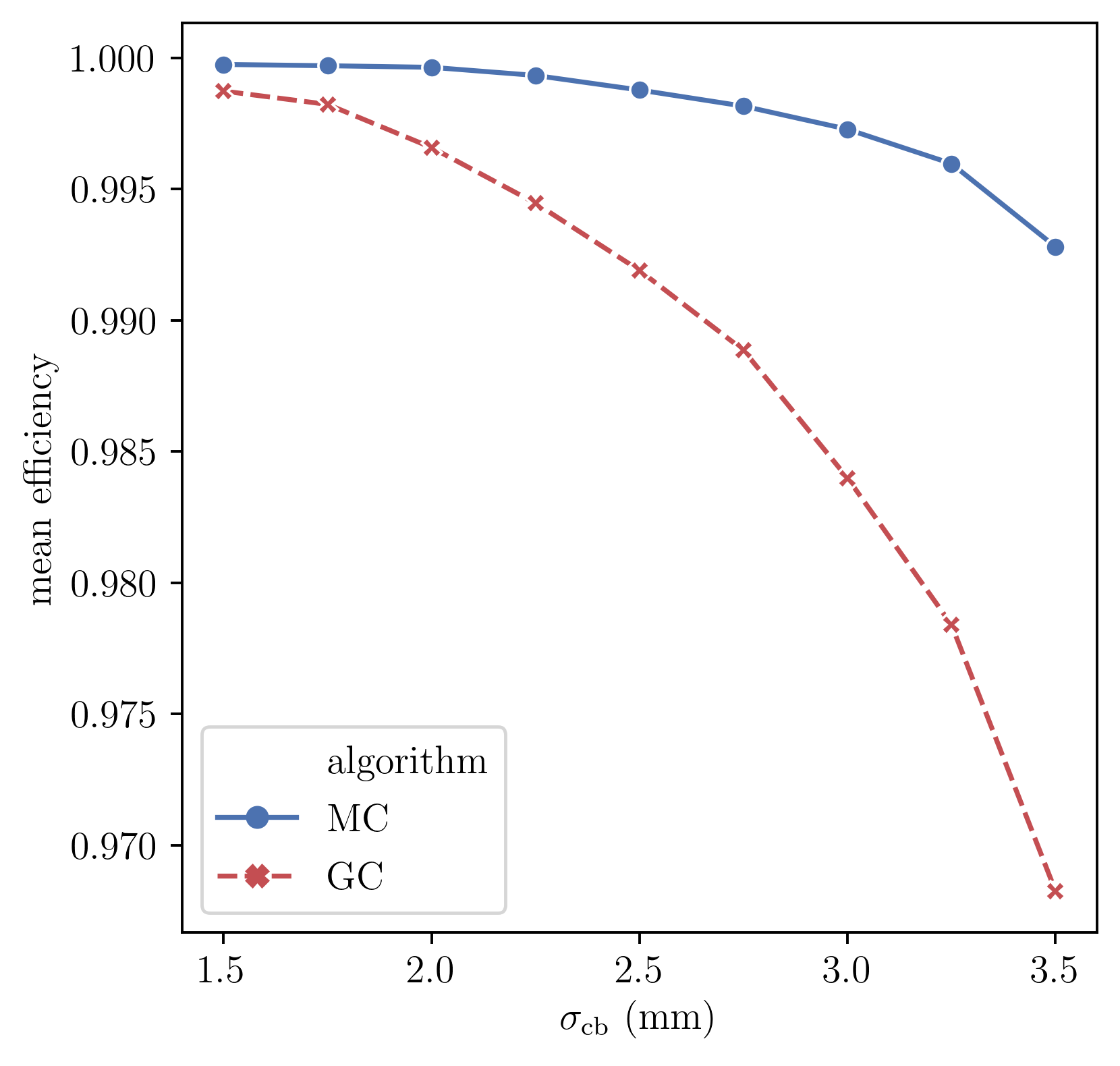}
\caption{Mean efficiency for MC (blue solid line) and GC (red dashed line) resulting from the suite of trials described in Table \ref{tab:sim1}.  GC mean efficiency is $>0.998$ for $\sigma_\mathrm{cb} < 2$ mm.  MC mean efficiency is $>0.998$ for $\sigma_\mathrm{cb} < 3$ mm.  The MC algorithm outperforms the GC algorithm over the full range of $\sigma_\mathrm{cb}$.  Both algorithms see monotonic decreases in efficiency as the level of crowding increases.  An efficiency of $0.998$ corresponds to a single source coordinate replacement in a grid of 547 positioners.
\label{fig:means}}
\end{figure}

Figure \ref{fig:means} shows the trend of trial-averaged efficiency versus $\sigma_\mathrm{cb}$ for the MC and GC algorithms.  The GC algorithm maintains a mean efficiency greater than 0.998 for $\sigma_\mathrm{cb} < 2$ mm, the MC algorithm maintains a mean efficiency is greater than 0.998 for $\sigma_\mathrm{cb} < 3$ mm.  An efficiency of 0.998 corresponds to a single replacement in a grid of 547 positioners.  With either routine only a fraction of a percent of astronomical targets will be lost in the small $\sigma_\mathrm{cb}$ regimes.

The MC algorithm outperforms the GC algorithm in terms of efficiency at all $\sigma_\mathrm{cb}$ values.  For both algorithms, efficiency declines monotonically with $\sigma_\mathrm{cb}$.  Under the highest crowding conditions ($\sigma_\mathrm{cb} = 3.5$ mm) mean efficiencies of 0.993 and 0.968 are measured for the MC and GC algorithms.  In an environment three times more crowded than the SDSS-V FPS design, we are limiting source replacement to less than a percent using the MC algorithm.  This result suggests that very crowded RFP arrays may be feasible from a collision avoidance standpoint.

\begin{figure}[ht!]
\plotone{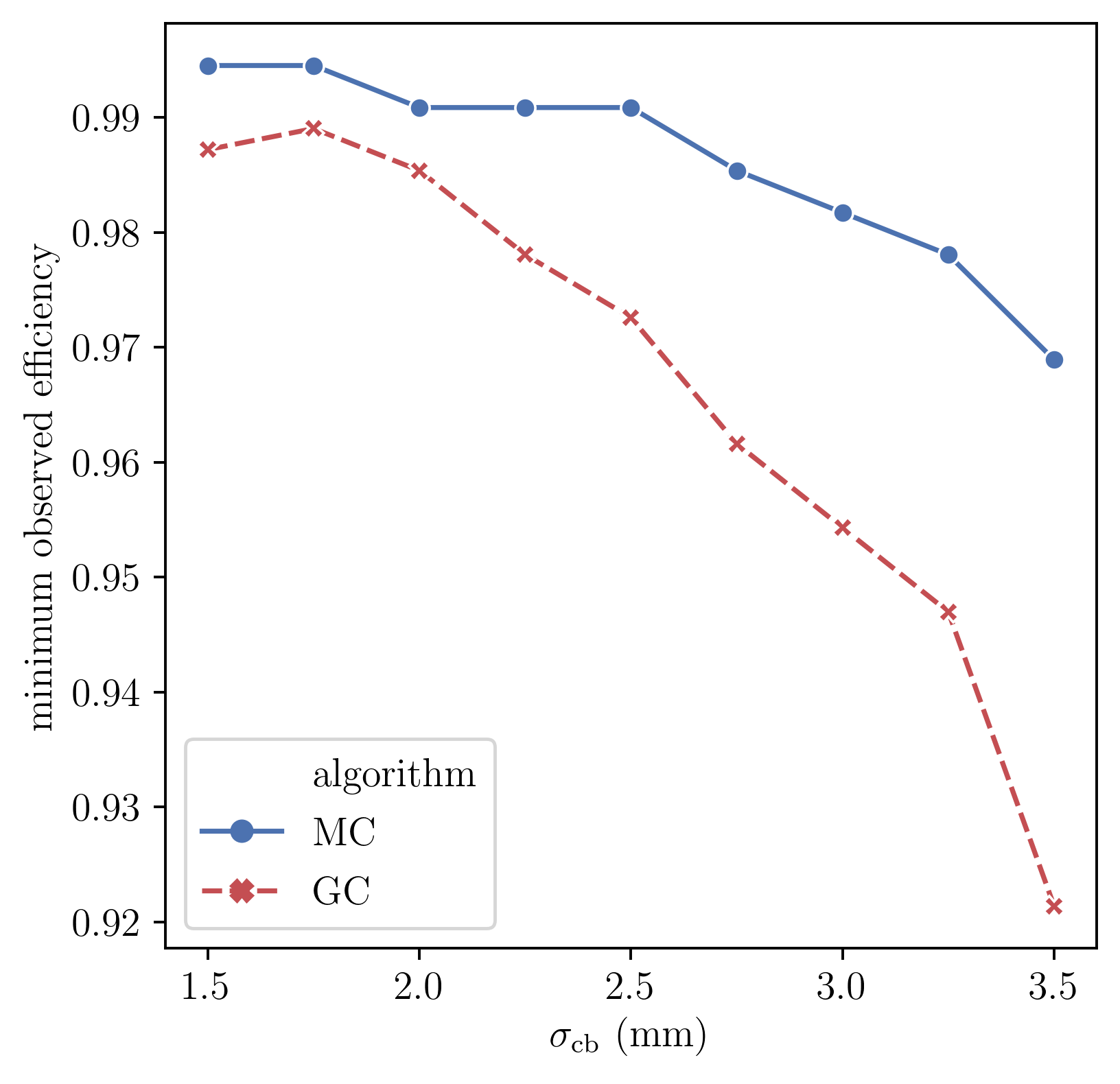}
\caption{Lowest observed efficiencies for MC (blue solid line) and GC (red dashed line) seen in the suite of 36,000 trials described in Table \ref{tab:sim1}.  For $\sigma_\mathrm{cb} \leq 2$ mm, the minimum observed efficiency is $>0.98$.  For $\sigma_\mathrm{cb} > 2$ mm, the minimum seen efficiency steadily drops, reaching a minimum around 0.92 for the GC algorithm and 0.97 for the MC algorithm.
\label{fig:mins}}
\end{figure}

Figure \ref{fig:mins} shows the minimum observed efficiency as a function of algorithm and $\sigma_\mathrm{cb}$ in the set of 36,000 trials.  The trends are similar for both algorithms.  For $\sigma_\mathrm{cb} \leq 2$ mm, minimum efficiency remains above 0.98, beyond $\sigma_\mathrm{cb} = 2$ mm, minimum efficiency trends steadily downward.  For the largest $\sigma_\mathrm{cb}$, we see minimum efficiencies around 0.92 for the GC algorithm and 0.97 for the MC algorithm.

\begin{figure}[ht!]
\plotone{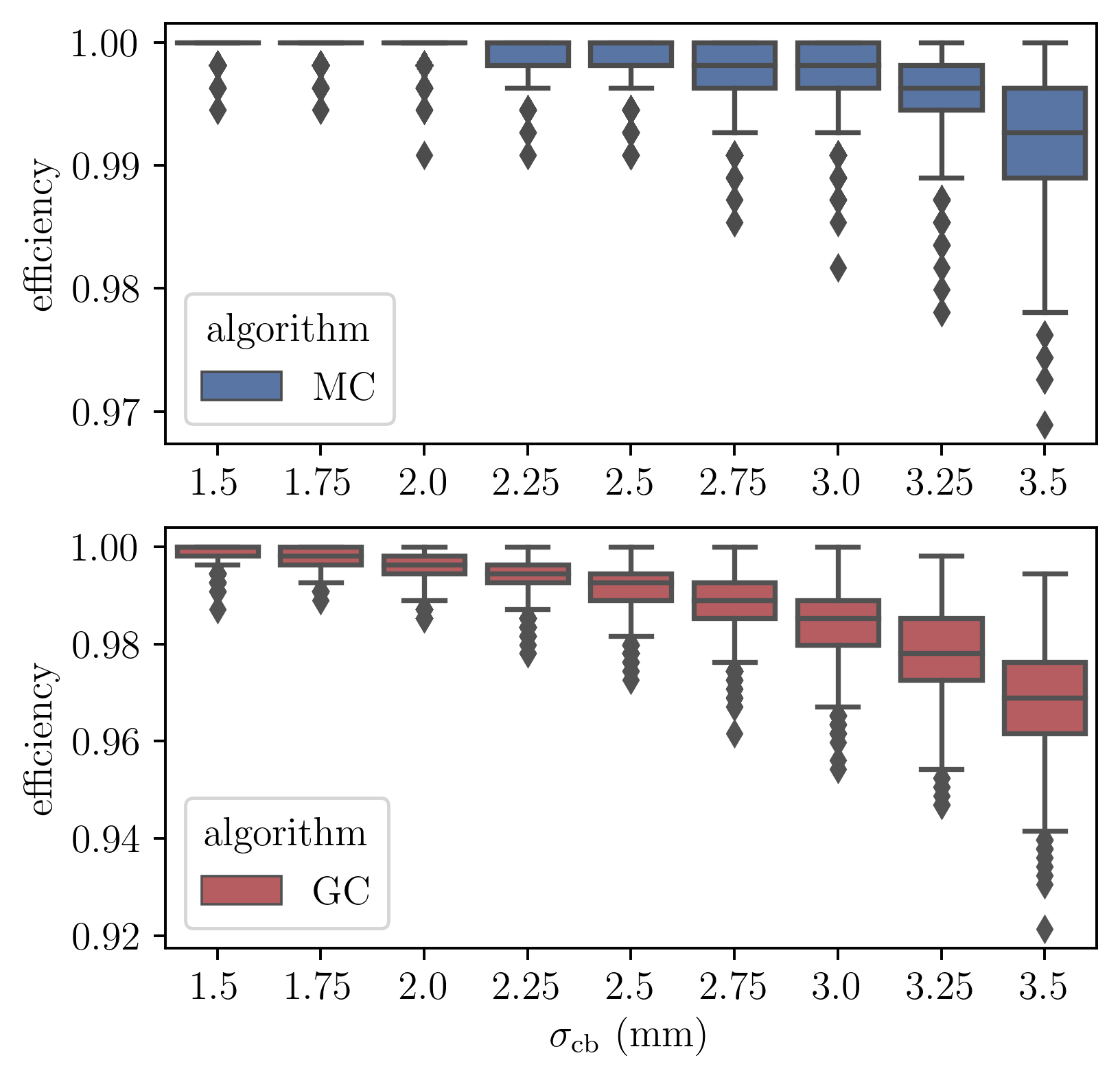}
\caption{Box plots of the underlying efficiency distributions for the MC (top panel) and GC (bottom panel) algorithms from the suite of trials described in Table \ref{tab:sim1}.  Boxes capture the innerquartile range, and whiskers capture the data lying within 3/2 of the innerquartile range below and above the low and high quartiles.  Points lying outside the whiskers are marked as diamonds.  With the MC algorithm, target loss will be limited to a small fraction of a percent over the whole $\sigma_\mathrm{cb}$ range.  Even in the most crowded regime, the MC algorithm will see a only a small target loss percentage of $\sim$1\%.  The GC algorithm limits target loss 1\% for the lower half of $\sigma_\mathrm{cb}$ values investigated.
\label{fig:box}}
\end{figure}

Figure \ref{fig:box} presents box plots of measured efficiencies for the MC (upper panel) and GC (lower panel) algorithms across trials.  Boxes indicate the innerquartile range, and whiskers capture data within 3/2 the inner quartile range below and above the low and high quartiles.  Outliers are plotted as diamonds.  Generally the variance of efficiency increases with increasing $\sigma_\mathrm{cb}$.  This figure indicates a median target loss on the sub-percent level for the MC algorithm over the full range of $\sigma_\mathrm{cb}$ investigated.  Sub-percent target loss for the GC algorithm is typical when $\sigma_\mathrm{cb} < 2.5$ mm.

\subsection{Reconfiguration Time}

Reconfiguration times for RFP arrays are usually measured in seconds.  However, when integrated over years of a survey, these seconds add up to significant hours of observational overheads.  Here we analyze the reconfiguration times from the simulation suite described in Table \ref{tab:sim1} using SDSS-V positioner velocities.  In this section we measure ``fold time", which is the time required to move between a source (or astronomical target) orientation to a folded state.  The total field-to-field reconfiguration time is twice this value, as the positioner array must move first to the folded state before moving to the next desired orientation.

\begin{figure}[ht!]
\plotone{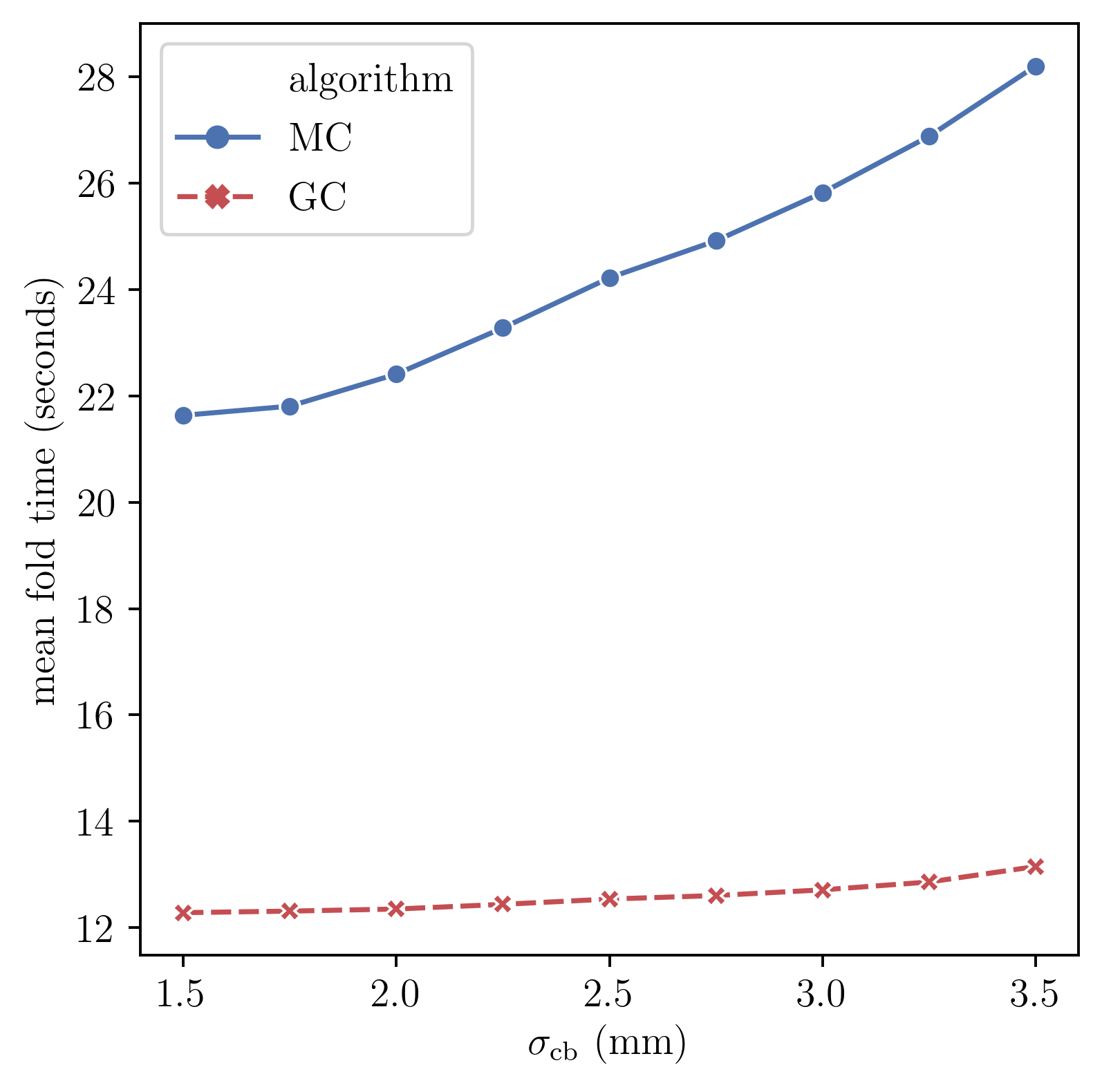}
\caption{Average time for an SDSS-V robot array to transition between a folded state and a source coordinate (on-target) state, representing half the time of a complete reconfiguration.  Results are compiled from the trials described in Table \ref{tab:sim1} for MC (blue solid line) and GC (red dashed line) algorithms.  The GC algorithm yeilds a total reconfiguration time less than 30 seconds.  The MC algorithm experiences longer reconfiguration durations due to the injection of stochastic motion along a robot's trajectory.  The amount of stochasticity is tunable, and thus the GC fold time curve represents a lower bound for the MC method dependent on parameter tuning.
\label{fig:meanFold}}
\end{figure}

Figure \ref{fig:meanFold} shows the mean fold time as a function of $\sigma_\mathrm{cb}$ and algorithm.  The GC outperforms the MC algorithm by a factor of $\sim$2, depending slightly on $\sigma_\mathrm{cb}$.  The longer path lengths observed in the MC algorithm are due to two effects: (1) the injection of random motion along its path, tuned by the greed parameter, and (2) the additional policy of $E_i$ (energy) minimization, tuned by the phobia parameter.  Recall that, as greed approaches 1 and phobia approaches 0, the MC algorithm becomes identical to the GC algorithm.  In this sense the fold time curve for the GC algorithm in Figure \ref{fig:meanFold} represents a lower bound for the MC algorithm.  By varying greed and phobia, one might tune the MC algorithm to produce an optimal balance between efficiency and fold time in the context of an astronomical survey.  Here we have fixed greed and phobia to 0.9 and 0.3 to provide a point comparison between the two methods.

\begin{figure}[ht!]
\plotone{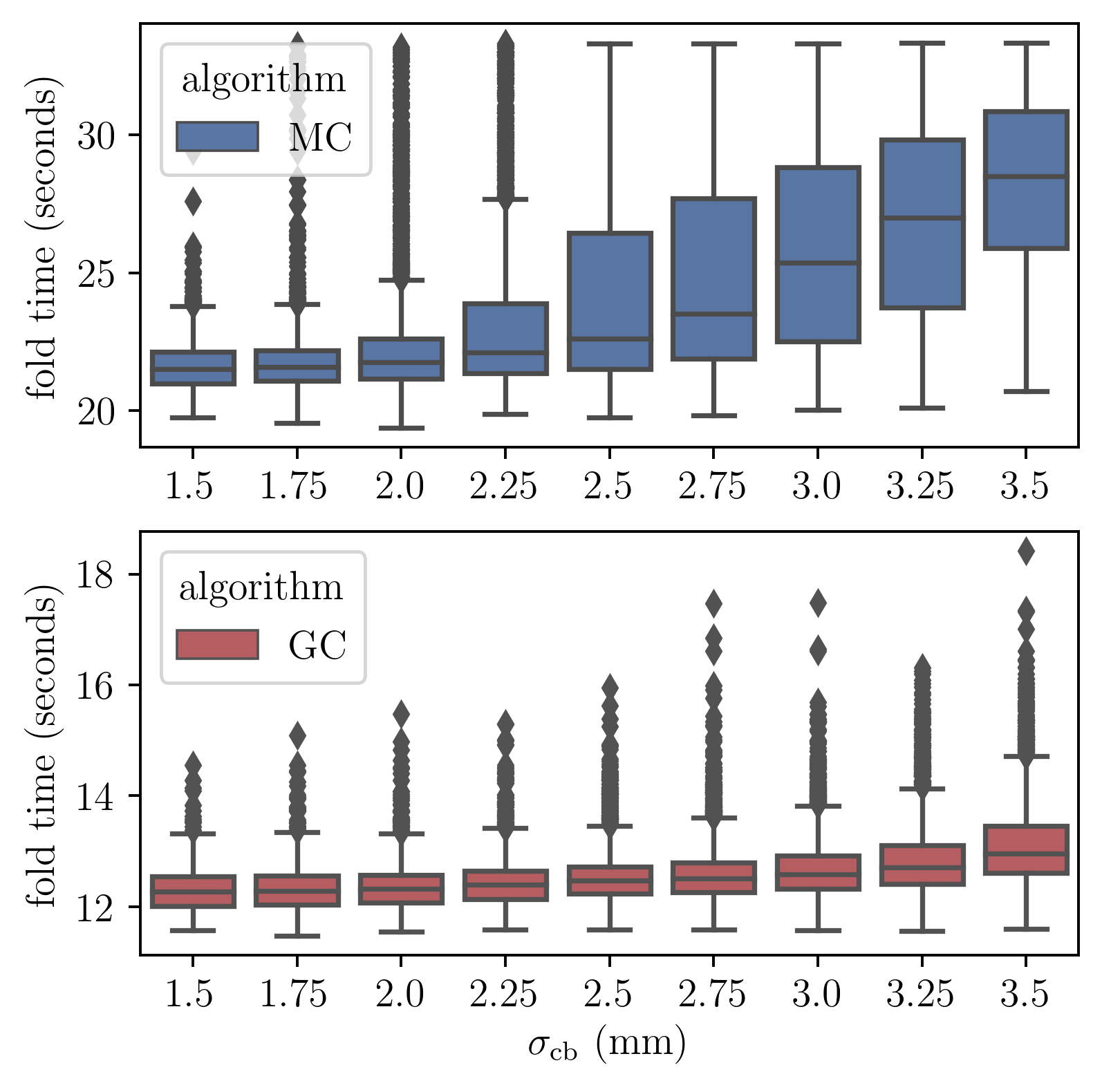}
\caption{Box plots for the distribution of fold times seen across the $\sigma_\mathrm{cb}$ range for the trials described in Table \ref{tab:sim1}.  Boxes capture the innerquartile range, and whiskers capture the data lying within 3/2 the inner quartile range below and above the low and high quartiles.  Data outside whiskers are plotted as diamonds.  The MC algorithm yeilds longer median fold times with a higher level of variance when compared to the GC algorithm.
\label{fig:boxFold}}
\end{figure}

Figure \ref{fig:boxFold} shows box plots for fold time across $\sigma_\mathrm{cb}$ and between algorithms. Boxes indicate the innerquartile range, and whiskers capture data within 3/2 the inner quartile range below and above the low and high quartiles.  Data falling outside whiskers are indicated with diamonds.  For the GC algorithm we see median reconfiguration times less than 30 seconds for SDSS-V positioners across the full range of $\sigma_\mathrm{cb}$.  For the MC algorithm, we expect reconfiguration times closer to 45 seconds for the lower half of $\sigma_\mathrm{cb}$ values, and reconfiguration times nearing a minute for the upper half of the $\sigma_\mathrm{cb}$ value range.  The GC algorithm generally achieves the SDSS-V benchmark goal of a 30 second RFP reconfiguration time, whereas the MC would require greed and phobia parameter adjustments to reach this benchmark.

\subsection{Runtime and Efficiency vs. Step Size}
Navigating through a moving crowd is more effective when one takes small steps in the right direction, rather than waiting for the opportunity of a big step to present itself.  This analogy holds true for the algorithms we've presented.  However, a small step comes at the price of an extended runtime.

A fast running routine is desirable for at least two reasons. A fast path generator allows for dynamic decision making throughout a night's observations, as optimal plans are subject to change on short notice.  Another benefit of a fast path generator is that it may be incorporated in the optimization stages of survey planning and design.  In a large astronomical survey, field assignment for millions of objects is a computationally intensive task, so if path verification is required at this stage it must have a low computational overhead.  Carrying out path planning during field assignment is desirable as it will inform which specific targets ultimately get observed, providing an opportunity for vetting long before a field is visited by the telescope.

We measure two distinct runtimes in our routines.  The first runtime is the source replacement runtime $\tau_\mathrm{sr}$.  This is the computation time accumulated during the source replacement procedure described in Section \ref{sec:replace}.  The source replacement procedure requires iterative runs of the path generator to find valid initial coordinates.  $\tau_\mathrm{sr}$ is largely an overhead during target assignment, as once initial coordinates have been assigned, and paths shown to converge, they do not need to be recomputed.

The second runtime $\tau_\mathrm{pg}$ is the computational time required to build paths from a set of initial coordinates that are known to converge.  Observing conditions during nightly operations may require that RFP trajectories remain flexible on short notice.  For example, the airmass of observation may require slight adjustments to a precomputed set of initial coordinates.  Or perhaps a robot malfunctions and must be taken off-line during the night while remaining in a position that obstructs neighbors.  In cases where we envision small adjustments to initial coordinates, we expect the overall runtime of path generation to be much closer to $\tau_\mathrm{pg}$ than $\tau_\mathrm{sr}$.

For this experiment we vary $\Delta_\theta$, $\sigma_\mathrm{cb}$, and algorithm.  We run 300 trials at each unique parameter combination, yeilding a total of 18000 trials.  An analysis of the resulting trends informs a smart decision on step size: a value that produces a reasonable balance between efficiency and runtime.  The parameters for this simulation are listed in Table \ref{tab:sim2}.

\begin{table}[h]
\caption{Simulation 2 - vaired crowding and step size} \label{tab:sim2}
\begin{tabular}{rl}
\hline\hline 
Parameter             & Values   \\
\hline
n robots & 547        \\
n trials \footnote{trials split evenly amongst the varied $\sigma_\mathrm{cb}$, $\Delta_\theta$, and algorithm}               & 18000      \\
$\mathrm{l}_\alpha$ (mm)                   & 7.4  \\
$\mathrm{l}_\beta$ (mm)                   & 15   \\
$\vecc{\theta}^\mathrm{\,I}$ (deg)                   & random, right-armed   \\
$\vecc{\theta}^\mathrm{\,D}$ (deg)                   & (10,170) \\
pitch (mm)                & $\mathrm{l}_\alpha + \mathrm{l}_\beta = 22.4$ mm     \\
$\Delta_\theta$ (deg)                   & $\{0.05, 0.1, 0.25, 0.5, 0.75, 1\}$        \\
max steps               & $1000^\mathrm{o}/\Delta_\theta$        \\
$\sigma_\mathrm{cb}$ (mm)                   & $\{1.5, 2, 2.5, 3, 3.5\}$        \\
greed                 & 0.9        \\
phobia                & 0.3 \\
algorithm & $\{\mathrm{GC}, \mathrm{MC} \}$\\
\hline
\end{tabular}
\end{table}

\begin{figure*}[ht!]
\epsscale{1}
\plotone{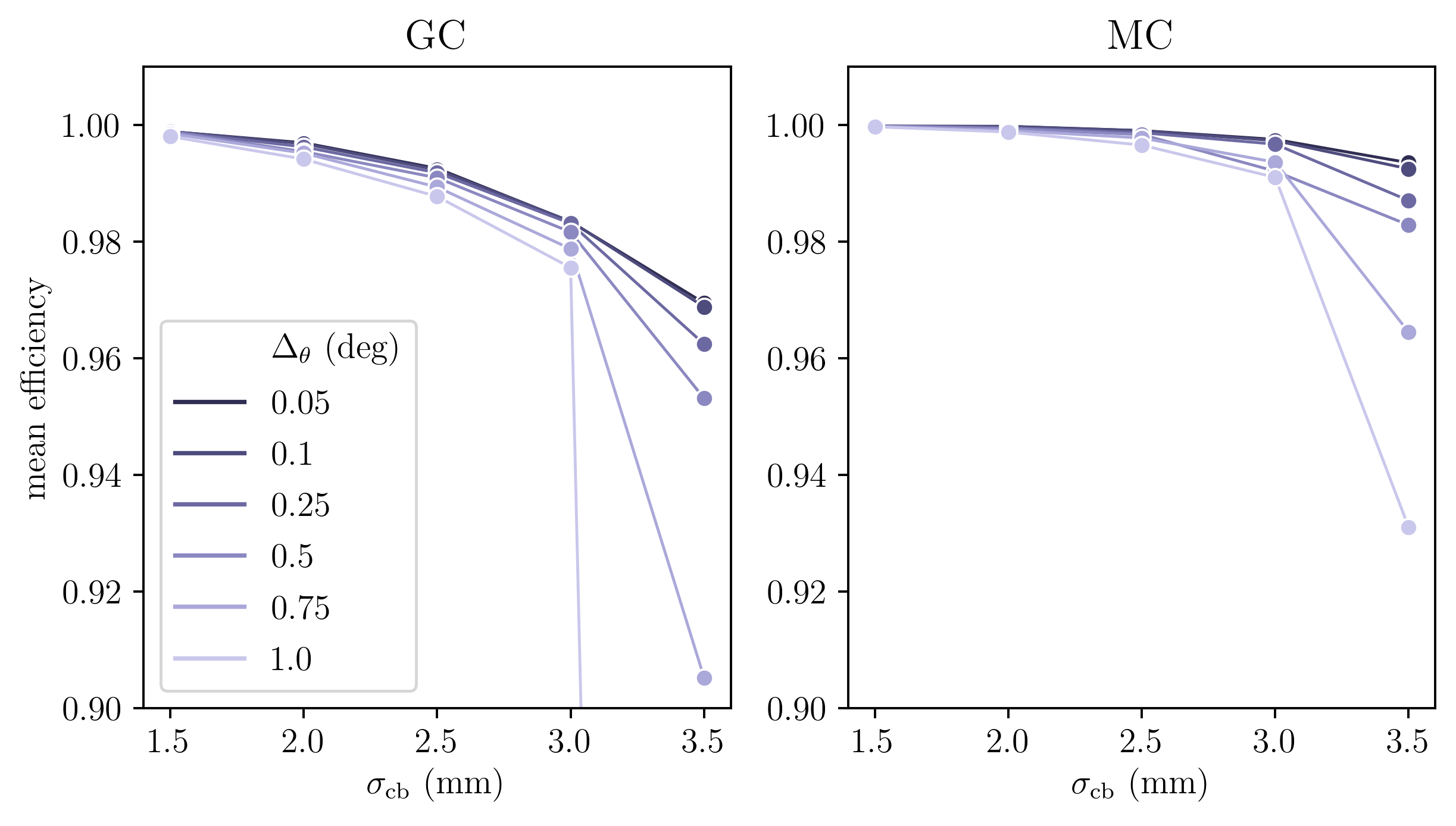}
\caption{Plots of efficiency for GC (left panel) and MC (right panel) algorithms over a range of crowding ($\sigma_\mathrm{cb}$) and step size ($\Delta_\theta$) settings, averaged across trials from the simulation set described in Table \ref{tab:sim2}.  Efficiency declines as step size is increased.  The combination of large step size and high crowding shows drastic reductions in efficiency.  In low crowding environments, choice of step size has only a minor effect on overall efficiency.  This figure ultimately illustrates the trade off between runtime and efficiency, as larger step sizes are preferable from a program runtime perspective.
\label{fig:meanEffVarStep}}
\end{figure*}

Figure \ref{fig:meanEffVarStep} shows the effects of $\Delta_\theta$ on efficiency for both GC and MC algorithms.  Efficiency declines as both $\Delta_\theta$ and $\sigma_\mathrm{cb}$ increase.  At small $\sigma_\mathrm{cb}$ (1.5-2 mm), there is little variation in efficiency, so large step sizes (0.75-1 deg) seem permissible.  However, as $\sigma_\mathrm{cb}$ increases, smaller step sizes are required to remain near the high end of efficiency.  This is especially true for $\sigma_\mathrm{cb} = 3.5$ mm, where efficiency is strongly influenced by step size.  The efficiency degradation with increased step size is due to the maximum displacement (MD) factor entering in Equation \ref{eq:4}, which defends against moves during which robots may undetectably jump through a colliding orientation in a single step.

\begin{figure*}[ht!]
\epsscale{1}
\plotone{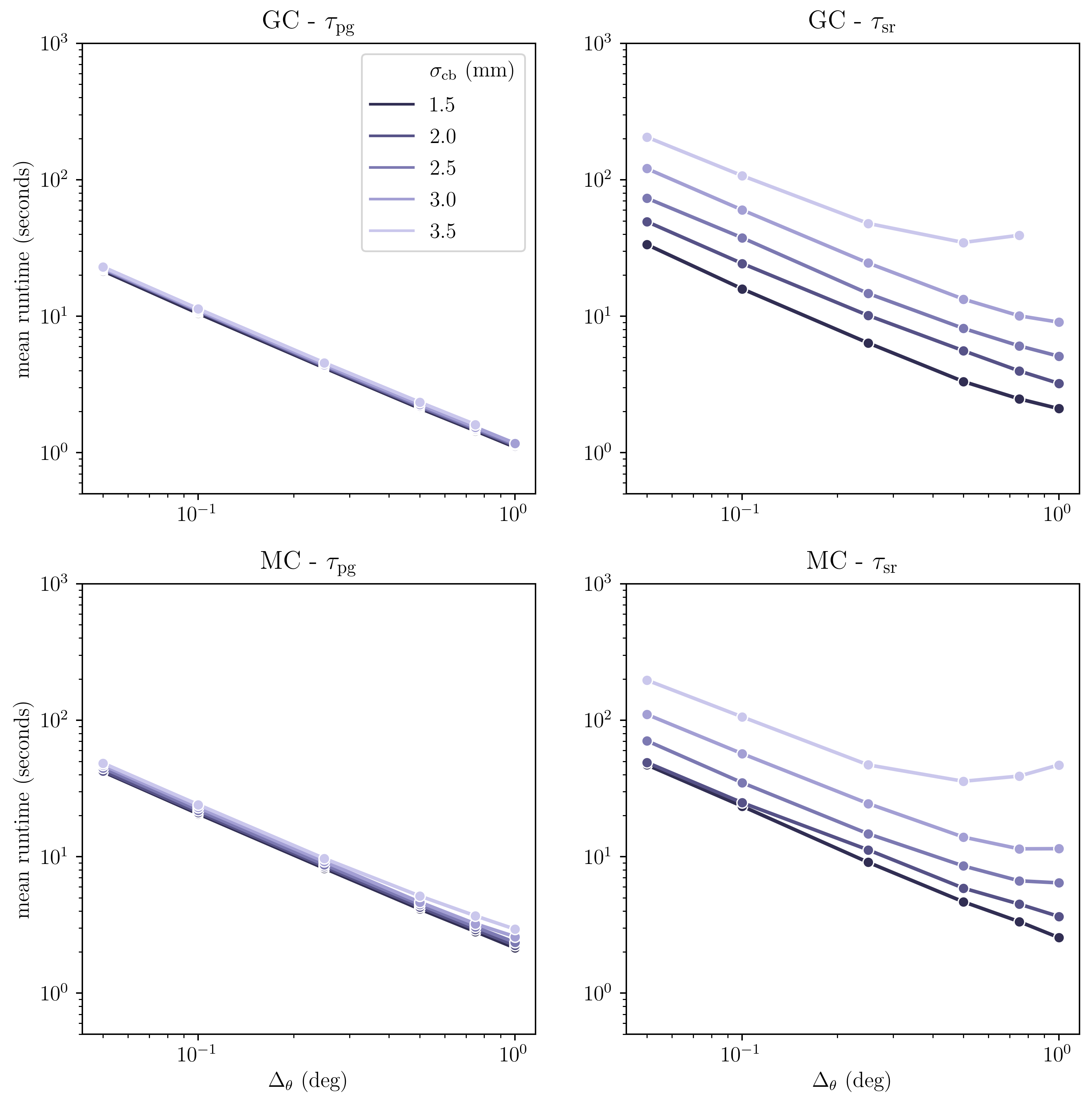}
\caption{Mean runtimes from the Table \ref{tab:sim2} simulation set.  Only results where mean efficiency $> 0.9$ are included.  The left two panels show the mean runtime for a single converging run of the path generator ($\tau_\mathrm{pg}$).  For the GC algorithm, a step size of 1 deg requires $\sim$1 second to compute paths for a grid of 547 positioners, a step size of 0.05 requires $\sim$20 seconds.  The MC algorithm requires roughly twice the runtime of the GC algorithm and exhibits a slight dependence on crowding.  The right two panels show the cumulative runtime $\tau_\mathrm{sr}$ which includes the iterative procedure of source coordinate replacement.  $\tau_\mathrm{sr}$ is strongly influenced by crowding.
\label{fig:meanRTVarStep}}
\end{figure*}

Figure \ref{fig:meanRTVarStep} presents the average observed runtimes across $\sigma_{cb}$, $\Delta_\theta$, and algorithm.  Here we only consider results for which the trial-averaged efficiency is greater than 0.9.  The two left hand panels show the average runtime $\tau_\mathrm{pg}$.  Over the full range of $\Delta_\theta$, the runtime decreases exponentially, ranging between $\sim 1-25$ seconds for the GC algorithm.  The MC algorithm runtime requires roughly twice the runtime of the GC algorithm, with a slight dependence on $\sigma_{cb}$.  The right hand panels of Figure \ref{fig:meanRTVarStep} show the average runtime $\tau_\mathrm{sr}$, which includes iterative process of source coordinate replacement (Section \ref{sec:replace}).  When including the source replacement procedure, the runtime scales by roughly an order of magnitude over the full range of $\sigma_\mathrm{cb}$.  This scaling is largely due to the decrease in overall efficiency as $\sigma_\mathrm{cb}$ increases.  As $\sigma_\mathrm{cb}$ increases, the chance of deadlock increases, and so the required number of iterations of the path generator increase.

The trends indicated in Figures \ref{fig:meanRTVarStep} and \ref{fig:meanEffVarStep} provide the information we need to select decent $\Delta_\theta$ settings for each algorithm over a range of $\sigma_\mathrm{cb}$.  Table \ref{tab:res} contains a summary of results for each algorithm under some reasonable parameter choices.  The table contains runtime, fold time, and efficiency metrics over the full range of $\sigma_\mathrm{cb}$, providing a concise summary of the performance of the algorithms we have tested over a range of parameter space.

\begin{deluxetable*}{ccccccc}
\tablecolumns{7}
\tablecaption{Selected Results from Simulation 2 \label{tab:res}}
\tablewidth{0pt}
\tablehead{
 \colhead{algorithm} &
 \colhead{$\sigma_\mathrm{cb}$ (mm)} &
 \colhead{$\Delta_\theta$ (deg)} &
 \colhead{mean $\tau_\mathrm{pg}$ (sec)\tablenotemark{\scriptsize{a}}} &
 \colhead{$\tau_\mathrm{sr}$ (sec)\tablenotemark{\scriptsize{b}}} &
 \colhead{mean fold time (sec)\tablenotemark{\scriptsize{c}}} &
 \colhead{mean efficiency}
 }
\startdata
GC & 1.50 & 1.00 & 1.10 & 2.10 & 12.32 & 0.9980 \\
GC & 2.50 & 0.25 & 4.34 & 14.67 & 12.55 & 0.9918 \\
GC & 3.50 & 0.10 & 11.32 & 107.11 & 13.17 & 0.9688 \\
MC & 1.50 & 1.00 & 2.14 & 2.55 & 22.94 & 0.9997 \\
MC & 2.50 & 0.25 & 8.74 & 14.71 & 24.29 & 0.9987 \\
MC & 3.50 & 0.10 & 24.04 & 105.82 & 27.90 & 0.9925 \\
\enddata
\tablenotetext{\scriptsize{a}}{runtime of single path generator pass}
\tablenotetext{\scriptsize{b}}{runtime of multiple path generator pass with source replacement}
\tablenotetext{\scriptsize{c}}{half of the expected duration of robot motion during reconfiguration}
\end{deluxetable*}

In nightly operations, SDSS-V FPS path computations $\tau_\mathrm{pg}$ will typically require only few seconds, even when choosing a conservative $\sigma_\mathrm{cb} = 2.5$ mm.  Should a sudden RFP reconfiguration be desired, the overhead due to path planning computations will be negligible when compared to the duration of robot motion.  The computational speed of these algorithms provides SDSS-V with a nimble observing system that suffers almost no additional overhead when unscheduled reconfigurations are requested at a moment's notice.

\subsection{Runtime and Efficiency vs. Grid Size}
In this section we provide a brief analysis of how algorithmic performance scales with number of positioners.  We use relatively few trials (6000), and choose a mid-range crowding level of $\sigma_\mathrm{cb} = 2.5$ mm.  The complete set of simulation parameters are described in Table \ref{tab:sim3}.  The results here will give a general feel for behavior in grids more massive than the typical 547 robot array we have focused on thus far.

\begin{table}[h]
\caption{Simulation 3 - varied grid size} \label{tab:sim3}
\begin{tabular}{rl}
\hline\hline 
Parameter             & Values   \\
\hline
n robots & $\{37, 91, 169,...,2269,2611,2977\}$        \\
n trials \footnote{trials split evenly amongst the varied grid sizes and algorithm type}               & 6000      \\
$\mathrm{l}_\alpha$ (mm)                   & 7.4   \\
$\mathrm{l}_\beta$ (mm)                   & 15   \\
$\vecc{\theta}^\mathrm{\,I}$ (deg)                   & random, right-armed   \\
$\vecc{\theta}^\mathrm{\,D}$ (deg)                   & (10,170) \\
pitch (mm)             & $\mathrm{l}_\alpha + \mathrm{l}_\beta = 22.4$    \\
$\Delta_\theta$ (deg)                   & 0.1        \\
max steps               & $1000^\mathrm{o}/\Delta_\theta$        \\
$\sigma_\mathrm{cb}$ (mm)     & 2.5        \\
greed                 & 0.9        \\
phobia                & 0.3 \\
algorithm & $\{\mathrm{GC}, \mathrm{MC} \}$ \\
\hline
\end{tabular}
\end{table}

\begin{figure}[ht!]
\plotone{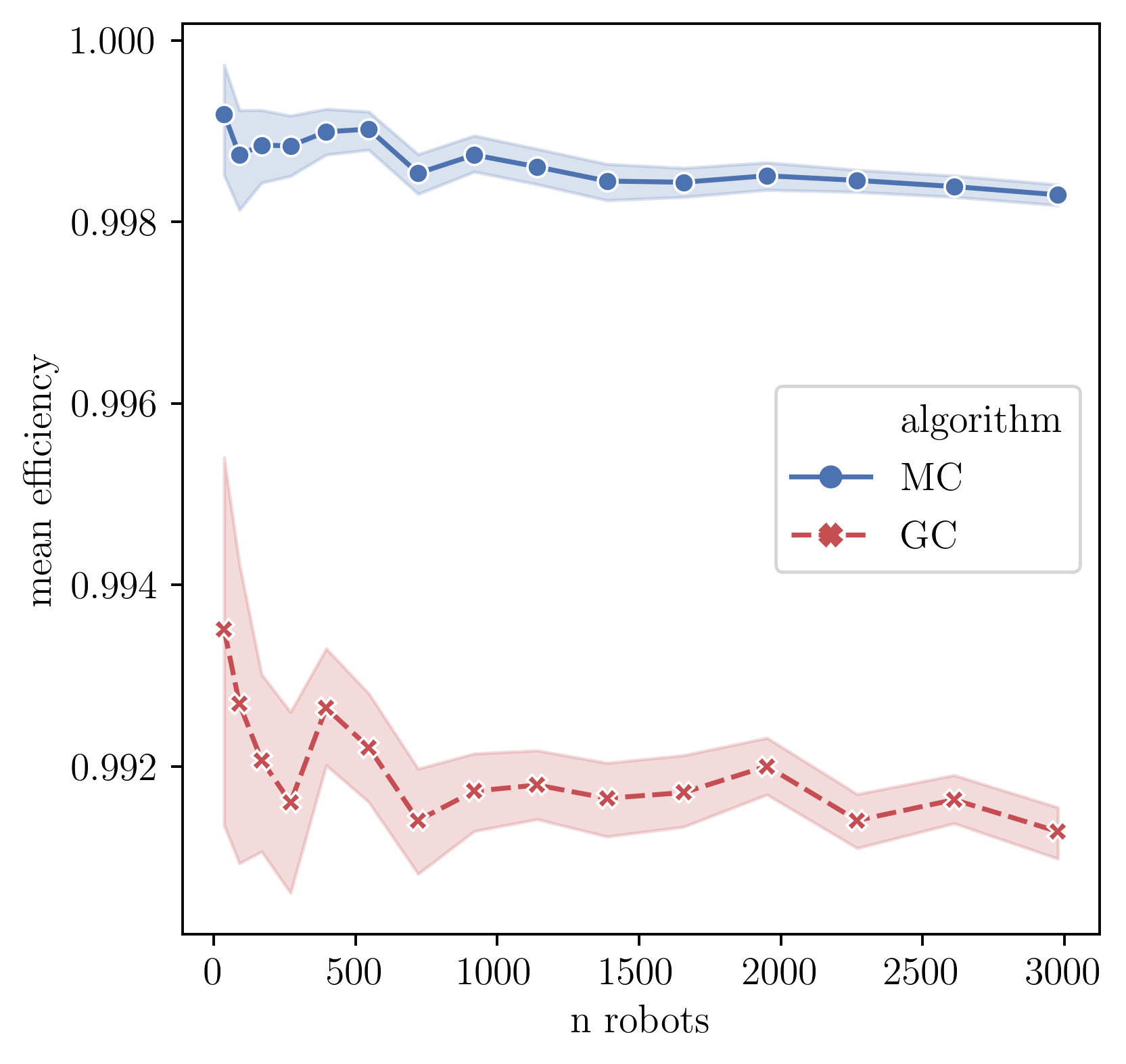}
\caption{A comparison of mean efficiency between MC and GC algorithms at various grid sizes using the parameters set in Table \ref{tab:sim3}.  Shaded regions indicate a 95\% confidence interval of the mean.  Efficiencies for smaller grids are computed from smaller samples of positioners, and thus are more variable than efficiencies measured from large grids.  For grids larger than SDSS-V's ($>500$ positioners), efficiency is not strongly affected by the total number of positioners in a grid.
\label{fig:nvsEff}}
\end{figure}

Figure \ref{fig:nvsEff} plots the mean efficiency against grid size for the simulations described in Table \ref{tab:sim3}.  The shaded region around each line represents the 95\% confidence interval of the mean.  For grids larger than the SDSS-V array, the efficiency remains relatively constant.  For smaller grids, a direct comparison of efficiency vs grid size should be taken with a grain of salt for two reasons: (1) smaller grids have a larger fraction of positioners without neighbors, and (2) efficiency is computed relative to the total number of positioners in a trial, where smaller grids contain fewer positioners.  Despite this, we see that efficiency is not strongly affected by total number of positioners.

\begin{figure}[ht!]
\plotone{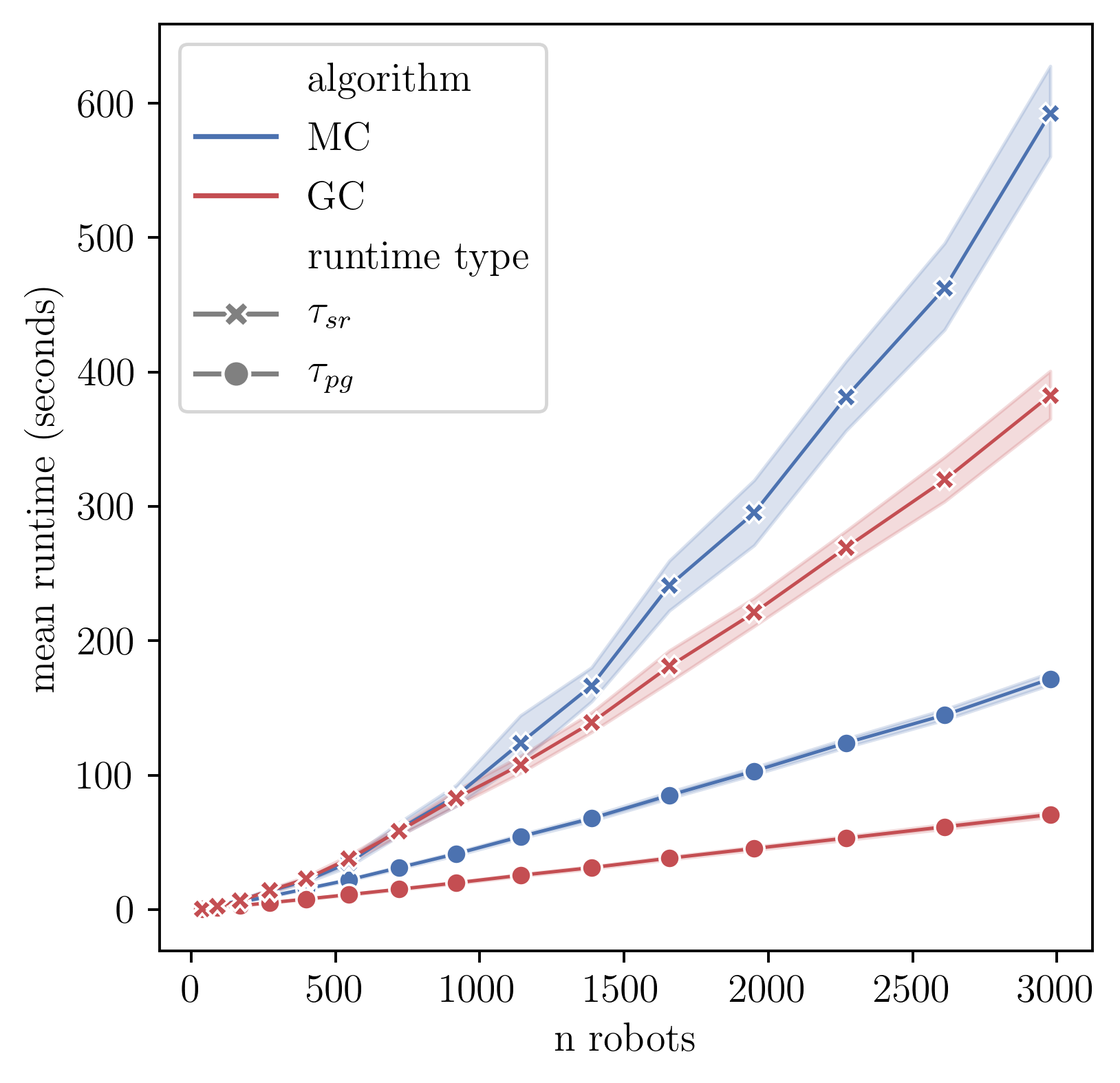}
\caption{A comparison of mean runtimes between MC and GC algorithms at various grid sizes using the parameters set in Table \ref{tab:sim3}.  Shaded regions indicate a 95\% confidence interval of the mean.
\label{fig:nvsRT}}
\end{figure}

Figure \ref{fig:nvsRT} plots the mean $\tau_\mathrm{sr}$ and $\tau_\mathrm{pg}$ runtimes for each algorithm at various grid sizes for the simulation set described in Table \ref{tab:sim3}.  For each line in the plot, a 95\% confidence interval of the mean is indicated by the shaded region.  Runtime $\tau_\mathrm{pg}$ increases linearly with grid size for both MC and GC algorithms, as is typical for distributed control algorithms.  Runtime $\tau_\mathrm{sr}$ does not show a tight linear response, as larger grids will experience a higher frequency of deadlock events, and thus require a higher number of source replacement iterations to solve a grid.


\section{Deployment Considerations} \label{sec:deployment}

We have made an effort to present generic routines that may be directly applied to many of today's RFP instruments via configurable parameters (pitch, l$_\alpha$, l$_\beta$, $\sigma_\mathrm{cb}$, etc).  To remain hardware-agnostic, we have inherently assumed grids of ``ideal" positioners, where the ideal positioner: (1) possesses infinite acceleration, (2) has sufficient on-board resources to follow an arbitrarily complex path, (3) has no positional uncertainty along a trajectory, and (4) never malfunctions.  When deploying the path generator in a realistic setting, points (1)-(4) above need to be considered carefully, and the optimal handling of these constraints may vary from one positioner design to another.  This section outlines specific strategies for addressing these issues for SDSS-V positioners, though we suspect they will be relevant for most other positioner designs and interfaces.

First we address points (1) and (2): a path post-processing strategy to account for the acceleration limits and finite on-board memory of an SDSS-V positioner.  The raw paths output by Kaiju intrinsically wiggle: when robot arms encounter each other they switch direction of motion frequently.  Applying a running-average filter to the velocity profiles for each axis sufficiently damps these wiggles to adhere to the acceleration limits of the hardware.  SDSS-V positioners accept a maximum of 1024 points per alpha or beta axis to define a trajectory, where a point is described by an (angle, time) tuple.  The embedded program of the SDSS-V positioner linearly interpolates between the supplied points.  We simplify the velocity-smoothed paths using the Ramer-Douglas-Peucker algorithm \citep{ramer}, which reduces the total number of points required to describe a path.  In the regime of small $\sigma_{\mathrm{cb}}$, a significant fraction of positioners are free to simply move at constant velocity to their destination without ever encountering a neighbor.  In these situations we find that many paths (especially GC paths with no random motion) are specified by only a handful of points.  This can be quickly realized from the left panel of Figure \ref{fig:cbuffrange}, where more than a few positioners have enough space to fold without interference.  As $\sigma_\mathrm{cb}$ increases, the probability of interactions increases, and the paths necessarily become more complex.  We are generally able to represent the most complex paths in less than 250 points after velocity-smoothing and point-simplification.  After a simplified path is computed, Kaiju is used to verify it (with a slightly reduced $\sigma_{\mathrm{cb}}$) to ensure that no collisions were created in the post-processing procedure.  As a consequence, path smoothing and simplification will require some amount of the overall $\sigma_\mathrm{cb}$ budget.

To address point (3) we must choose $\sigma_{\mathrm{cb}}$ large enough to account for the uncertainty of a robot's position along a path.  The dominant source of positional uncertainty is due to slight imperfections in the manufacture and assembly of a robot's gearbox.  This manifests as a nonlinear relationship between a positioner's commanded angular velocity and the true angular velocity, where a positioner's true velocity will oscillate about the expected velocity by a small amount in a repeatable way.  \cite{luzius} provide a comprehensive analysis of all the various sources of mechanical error in an SDSS-V positioner.

Collecting points (1)-(3) the selection of $\sigma_{\mathrm{cb}}$ must be the summation of three factors:

\begin{equation}
\sigma_\mathrm{cb} = \sigma_\mathrm{arm} + \sigma_\mathrm{smooth} + \sigma_\epsilon
\end{equation}

$\sigma_\mathrm{arm}$ is the half-width of the beta arm (1.5 mm for SDSS-V).  $\sigma_\mathrm{smooth}$ is an extra buffer required for path smoothing and simplification.  We've found that $\sigma_\mathrm{smooth} \sim 0.03$ mm is sufficient for ensuring collision-free trajectories after path smoothing at small step sizes ($\Delta\theta \sim 0.1$).  $\sigma_\epsilon$ is the maximum lateral uncertainty of the beta arm's absolute position due to mechanical errors.  Algorithmic performance increases as $\sigma_{\mathrm{cb}}$ shrinks, so this will encourage finding tight bounds on $\sigma_\epsilon$ for an SDSS-V RFP.

An a priori choice of $\sigma_{\epsilon}$ can be informed directly from the SDSS-V positioner requirements where bounds on four independent sources of angular error are specified: nonlinearity, hysteresis, fiber torque, and dynamic control error.  When these bounds are constructively summed the maximum angular error is $\pm 1.50$ and $\pm 1.48$ deg for the alpha and beta axes.  $\sigma_\epsilon$ will be maximized at the location of the fiber (at the end of beta arm) when the positioner is at full extension ($\theta_\beta=0$).  A simple computation for $\sigma_\epsilon$ can be constructed using Equation \ref{eq:1}.  For simplicity, take the positioner's base position to be $\vecc{\mathrm{b}} = (0,0)$.  Take $\vecc{\theta}_1=(0,0)$ deg to be the expected angular coordinates of a robot at full extension, with corresponding Cartesian fiber coordinates $\vecc{\mathrm{f}}_1 = (0, 22.4)$ mm.  Take $\vecc{\theta}_2 = (1.50, 1.48)$ deg to be the actual angular coordinates after the maximum angular uncertainties have been applied.  The corresponding Cartesian fiber coordinates are $\vecc{\mathrm{f}}_2=(22.38, 0.97)$ mm.  The resulting lateral uncertainty can be estimated as:

\begin{equation} \label{eq:tol}
\sigma_\epsilon = \| \vecc{\mathrm{f}}_1 - \vecc{\mathrm{f}}_2 \| = 0.97 \,\mathrm{mm}
\end{equation}

Thus, a setting of $\sigma_\mathrm{cb} = 1.5 + 0.03 + 0.97 = 2.5$ mm would be selected based on the summation of toleranced errors from SDSS-V positioner requirements.

Note that the calculation of $\sigma_\epsilon$ by Equation \ref{eq:tol} represents a ``worst case scenario" in which all positioner errors are maximized and summed constructively, and this situation is unlikely to happen.  SDSS-V robots are in active production and a comprehensive calibration and quality assurance procedure is in place for each positioner.  The QA process provides accurate measurements of positioner error as a function of angular coordinates and direction of motion.  Initial results show that positioners are significantly exceeding specifications, and the nonlinear and hysteresis effects are highly repeatable.  By modeling nonlinearity and hysteresis, we can more accurately predict the absolute orientation of a beta arm.  Based on initial calibration results we expect that a setting of $\sigma_\mathrm{cb}=2$ mm will be achievable.  Ultimately, SDSS-V's selection for $\sigma_\mathrm{cb}$ will be assessed after the full characterization for each positioner is received.  We will likely begin operations with a larger than necessary $\sigma_\mathrm{cb}$, to ensure protection against unexpected sources of positional error.

Finally, we comment on point (4): the case of a malfunctioning robot.  Throughout the duration of a multi-year survey operating with 1000 robots distributed between two hemispheres, it's plausible to assume that a robot may be required to be taken off-line for a period of time while remaining as fixed interfering obstacle within the grid.  The anti-collision routine we have presented can model this situation by simply setting the greed parameter to zero for any broken positioner in any orientation.  Operational robots will still be required to complete successful fold and unfold sequences, but any off-line robot will remain fixed between subsequent target orientations.  In this sense, it will be the duty of neighbors to navigate around a static obstacle.  A robot stuck in an unfortunate (e.g. outstretched) orientation will increase the probability of deadlock in its immediate area.  In certain orientations, a stationary robot could require additional neighboring robots to be taken off-line.  A robot fixed in a folded orientation will have little adverse effect on path generation.  In summary, operations may proceed even in the case of stationary robots using the anti-collision algorithm we have presented, although lower efficiencies would be expected.

The complete path generation and post-processing procedure has been successfully applied to SDSS-V positioners in a lab setting, and a video demonstration of reconfiguration tests is available online\footnote{\url{https://www.youtube.com/watch?v=jV3Uz_C8W0c}}.  A $\sigma_{\mathrm{cb}}=2$ mm setting was used in this video demonstration.

\section{Discussion} \label{sec:discussion}

In addition to path routing, Kaiju is an important component of higher-level algorithms that seek to globally optimize SDSS-V survey planning, design, and overall strategy.  Note that Kaiju possess no notion of concepts like target priority.  In SDSS-V certain targets (e.g. calibration sources) may be more important than others in a field.  In this work we have only presented a very simple deadlock resolution strategy that iteratively and randomly tosses away targets until a converging grid is achieved.  In practice, target assignment algorithms will preferentially select targets to throwaway only after searching for viable target replacements first.  In a grid of 500 heavily overlapping positioners, many opportunities for target swaps between positioners exist, which may serve to eliminate either source coordinate collisions or deadlocked pairs with no loss of original targets.  Furthermore, deadlocks may be rectified via focused manipulation of an individual robot's greed and phobia parameters in repeated trials.  The options are vast.  The targeting optimization challenge is arguably at least as complex as the problem of safe path generation, and it's an active area of algorithmic development for SDSS-V that will likely see continuous improvement throughout the survey.  In the hierarchy we describe, the selective mitigation of target attrition due to collision avoidance constraints is expected to be delegated to a higher-level piece of software with an awareness of the bigger picture.  The fast runtime and high efficiency of Kaiju make this division of duty computationally feasible, especially in the small $\sigma_{\mathrm{cb}}$ regime.

SDSS-V fields span a wide range of target densities, though the majority of fields lie in the densely populated Galactic plane.  In dense fields, we plan to use the GC algorithm, which provides maximal reconfiguration speed and minimal runtime.  Though the GC algorithm is slightly less efficient, dense fields have a surplus of targets from which to select suitable replacements.  For sparser fields, where target replacement opportunities are less abundant, we plan to use the MC algorithm to maximize targeting efficiency.  With the options provided between the MC and GC algorithms, we may balance runtime, efficiency, and reconfiguration time to maximize survey productivity over the whole sky.  We expect to see overall targeting efficiencies of $>$99.9\% and average reconfiguration times $<$ 30 seconds for SDSS-V.  For SDSS-V, a 30 second reconfiguration time is comfortably less than the readout duration for the BOSS spectrograph.  With target transitioning happening during readout, robot reconfiguration should introduce no extra overhead in the nightly observing sequence.

For projects outside the SDSS-V context, a quicker reconfiguration time may be important.  Our reconfiguration process requires that every RFP array transition routes through a common folded state, where the folded state is near the edge of travel for each positioner.  In this case, the total distance traveled by a positioner between targets will usually be large when compared to the distance between the subsequent targets.  \cite{2014A&A...566A..84M} modeled direct transitions between random positioner configurations with good results in slightly overlapping RFP workspace regimes.  In heavily overlapping RFP regimes, \cite{matin} comment on the challenge of fitting the durations of both path computation and robot reconfiguration within timescales that do not introduce significant operational overheads.

The common state transition we present does provide benefits, as it decouples the dependency of robot trajectory planning from the nightly observing sequence. As all paths are built between a target state and a common state, transitioning between any two target states becomes trivial.  This permits all paths to be computed and vetted during the survey design phases well ahead of observations.  For SDSS-V, the runtimes we experience are suitable for on-the-fly path generation, should the need arise.  However, for a theoretically large and crowded positioner array, a completely off-line path generator could prove necessary, and thus a common state transition would be mandatory to ensure that program runtime is not a hindrance to observing cadence.

Improvements in reconfiguration time might be found by allowing robot movement to end at a step prior to a completely folded state.  In the SDSS-V layout, collision-less motion is guaranteed while all positioners maintain $\theta_\beta \gtrsim 155$ deg, where this minimum value of $\theta_\beta$ is a level of folding that brings the beta arm completely within a radius of half the pitch.  The exact condition is given by:

\begin{equation} \label{eq:betamin}
\theta_\beta > 180^\mathrm{o} - \cos^{-1}\frac{(\frac{1}{2}\mathrm{pitch})^2 - \mathrm{l}^2_\alpha - (\mathrm{l}_\beta+\sigma_\mathrm{cb})^2}{2\mathrm{l}_\alpha\mathrm{l}_\beta}
\end{equation}

RFP arrays with small overlap zones would benefit most from an earlier exit criterion like Equation \ref{eq:betamin}, as beta arms need only to retract a small amount before finding themselves in a guaranteed collision-less environment.  Although shortcuts between target transitions for SDSS-V exist, the complete fold/unfold sequence we have presented is sufficiently fast to not rate-limit SDSS-V survey pacing.



It has become increasingly popular to retrofit existing telescopes with robotically filled focal plane arrays.  In the case of SDSS-V, positioner density was ultimately limited by the fiber capacity of the existing spectrographs.  In proposed future RFP projects with purpose-built spectrographs (e.g MegaMapper \citealt{megamapper}), the potential multiplexing power may be limited only by the density of positioners in the focal plane.  As RFP technology continues to progress, the possibility of ultra-densely packed arrays may be on the horizon.  In our analysis, we have shown high efficiencies ($>$0.99 for MC) in heavily-overlapping and highly-crowded arrays, where as much as 20\% of the focal plane space may be occupied by interfering robot arms (right panel, Figure \ref{fig:cbuffrange}).  The algorithms we present can be used to inform and verify layout choices in future RFP instrument design.  Most importantly, our simulations suggest that ultra-densely packed positioner layouts may be operationally feasible.

The success of our collision-avoidance method is mainly attributed to the realization that reverse-solved paths are extremely efficient.  We show that even an algorithm based on a simple greedy heuristic obtains an efficient solution to a challenging problem.  It may seem suspicious that the directionality should matter at all.  However, the reverse-solve strategy resembles a situation that arises in multi-agent pattern formation theory.  Multi-agent pattern formation problems are often concerned with deriving control laws that drive agents from random states to lattice-like structures and determining where convergence can be guaranteed \citep{flock}.  By adopting a reverse-solve direction for RFP arrays, we frame our problem in the same way, where the initial (target) state is random, and the final (folded) state is a lattice.  A lattice configuration only requires that all positioners share the same $(\theta_\alpha, \theta_\beta)$ coordinates.  Here we have specifically selected the folded lattice configuration for two reasons.  The first reason: as positioners fold they are driving toward a guaranteed collision-free environment (see Equation \ref{eq:betamin}), and the chance of interaction decreases with program step.  The second reason: throughout the routine $\theta_\alpha$ will decrease and $\theta_\beta$ will increase as a positioner moves from a right arm orientation toward a fold.  When alpha and beta axes move in opposite directions, the overall motion of the beta arm resembles a thrust rather than a swipe or a swing.  Given the elongated shape of a beta arm, a thrust presents a smaller cross section for interaction during motion.  We suspect that other existing RFP control codes will improve in efficiency if a reverse solution is employed, especially in heavily overlapping regimes.

The reverse-solve method is obviously beneficial when considering the geometries and kinematics specific to RFP arrays, though a cursory search of the robotics literature at large yields sparse mention of reversely solved paths.  In one example, \citealt{surgery} find a reverse-path strategy to optimize the routing problem faced in minimally invasive surgeries.  We wonder if the concept of path-direction preference is generally extendable to path planning problems in the context of robotic control theory and multi-agent optimization.

The landscape of survey-based astronomy is progressing rapidly, largely due to the continuing development of highly-automated telescope and instrument systems.  As the complexity of observing systems continues to increase, overall survey productivity may become limited by our ability to design and implement effective control algorithms.  This was realized in SDSS-V, where the effective utilization of the full suite of SDSS-V positioners was an unsolved problem prior to this work.  Ultimately, robotic path routing occupies a small part of a larger picture of overall survey control.  Optimization of target assignment, survey scheduling, and feedback from data acquisition are all spaces in which further algorithmic development may have significant effects on the overall productivity of current and future astronomical surveys.

\section{Conclusion} \label{sec:conclusion}

We have presented a generic, computationally fast, and highly efficient method to determine non-colliding paths for a two armed robotic fiber positioner system, where we measure efficiency in terms of astronomical targets ultimately acquired under collision avoidance constraints.  The RFP design we describe has become typical in today's growing number of robotic wide-field multi-object spectroscopic instruments, and so our methods are applicable to a wide range of astronomical surveys and instruments.  We have focused our analysis on a layout in which robot arms have heavily overlapping patrol zones, a design currently unique to the SDSS-V and MOONS instruments for which the collision-avoidance problem is especially challenging.  We find that efficiency remains high even in environments significantly more crowded than the SDSS-V layout, suggesting the feasibility of ultra-densely packed positioner arrays in future RFP designs.

\section{Acknowledgments}

The authors thank Russell Owen for advice on software design and implementation, Mehran Mesbahi for discussions and materials related to multi-agent control, Rishi Pahuja \& Kal Kadlec for aid in solid model renderings, and Dan Sunday for his freely available geometric algorithms.  The authors thank the autonomous referee who's comments and suggestions greatly improved this work.

Funding for the Sloan Digital Sky Survey V has been provided by the Alfred P. Sloan Foundation, the Heising-Simons Foundation, and the Participating Institutions. SDSS acknowledges support and resources from the Center for High-Performance Computing at the University of Utah. The SDSS web site is www.sdss.org.

SDSS is managed by the Astrophysical Research Consortium for the Participating Institutions of the SDSS Collaboration, including the Carnegie Institution for Science, the Chilean Participation Group, the Gotham Participation Group, Harvard University, The Johns Hopkins University, L'Ecole polytechnique f\'{e}d\'{e}rale de Lausanne (EPFL), Leibniz-Institut f\"{u}r Astrophysik Potsdam (AIP), Max-Planck-Institut f\"{u}r Astronomie (MPIA Heidelberg), Max-Planck-Institut f\"{u}r Extraterrestrische Physik (MPE), Nanjing University, National Astronomical Observatories of China (NAOC), New Mexico State University, The Ohio State University, Pennsylvania State University, Space Telescope Science Institute (STScI), the Stellar Astrophysics Participation Group, Universidad Nacional Aut\'{o}noma de M\'{e}xico, University of Arizona, University of Colorado Boulder, University of Illinois at Urbana-Champaign, University of Toronto, University of Utah, University of Virginia, and Yale University.

\appendix
\section{Pseudocode Implementation} \label{sec:code}

A pseudocode implementation of the anti-collision algorithm to generate a reverse path for an RFP array follows below, favoring clarity over computational efficiency.  The routine assumes that a few functions are defined.  \textbf{dot(v1, v2)} and \textbf{norm(v1)} return the dot product and Euclidean norm for input vectors \textbf{v1, v2}.  \textbf{rand()} returns a uniform random value between 0 and 1.  \textbf{shuffle(array)} randomizes the order of elements in the input array. Vector subtraction and scalar multiplication is defined in the normal linear algebra sense.
\begin{lstlisting}
# -----------------
# runtime parameters
# -----------------

# note that setting greed=1 and phobia=0 results in the GC algorithm

alphaArmLen $\gets$ length of alpha arm, $\mathrm{l}_\alpha$
betaArmLen $\gets$ length of beta arm, $\mathrm{l}_\beta$
collisionBuffer $\gets$ buffer that describes the collision envelope, $\sigma_\mathrm{cb}$
dTheta $\gets$ maximum angular step size for a positioner axis perturbation, $\Delta_\theta$
maxIter $\gets$ maximum iterations to perform
greed $\gets$ desired greed setting between 0 and 1
phobia $\gets$ desired phobia setting between 0 and 1
maxDispl $\gets$ (alphaArmLen + betaArmLen)*sin(2*dTheta)  # Equation 7

# build the list of axis perturbations from dTheta
perturbArray $\gets$ []
foreach dAlpha in [-dTheta, 0, dTheta]:
  foreach dBeta in [-dTheta, 0, dTheta]:
    perturbArray.append([dAlpha, dBeta])


# ---------------------
# structure definitions
# ---------------------

struct Robot:
  # holds relevant coordinates and neighbors for a given positioner
  x $\gets$ x coordinate of positioner base position, $x_\mathrm{b}$
  y $\gets$ y coordinate of positioner base position, $y_\mathrm{b}$
  alphaCurr $\gets$ current alpha angle,  $\vecc{\theta}^\mathrm{\,C}_\alpha$
  betaCurr $\gets$ current beta angle,  $\vecc{\theta}^\mathrm{\,C}_\beta$
  alphaDest $\gets$ 10  # alpha angle destination,  $\vecc{\theta}^\mathrm{\,D}_\alpha$
  betaDest $\gets$ 170  # beta angle destination,  $\vecc{\theta}^\mathrm{\,D}_\beta$
  neighbors $\gets$ list of Robot instances with whom I risk collision (Equation 4)

struct Vector3:
  # simply, a 3-vector
  x $\gets$ x coordinate
  y $\gets$ y coordinate
  z $\gets$ 0 # work in the z=0 focal plane

struct LineSegment:
  # a container representing a line segment between two Vector3's v0 and v1
  v0 $\gets$ an instance of Vector3
  v1 $\gets$ an instance of Vector3


# --------------------
# function definitions
# --------------------

function dist3D_Line_to_Line(L1, L2):
  # Almost verbatim from Dan Sunday's algorithm: http://geomalgorithms.com/a07-_distance.html
  # this computes $D_{ij}$ used Equations 3, 6, and 9.
  parameters: L1, L2 each a LineSegment
  output: the minimum distance between L1 and L2.

  SMALL_NUM $\gets$ 0.00000001  # anything that avoids division overflow
  u $\gets$ L1.v1 - L1.v0
  v $\gets$ L2.v1 - L2.v0
  w $\gets$ L1.v0 - L2.v0
  a $\gets$ dot(u,u)         # always >= 0
  b $\gets$ dot(u,v)
  c $\gets$ dot(v,v)         # always >= 0
  d $\gets$ dot(u,w)
  e $\gets$ dot(v,w)
  D $\gets$ a*c - b*b        # always >= 0

  # compute the line parameters of the two closest points
  if (D < SMALL_NUM):          # the lines are almost parallel
    sc $\gets$ 0.0
    if (b > c):
      tc $\gets$ d / b
    else:
      tc $\gets$ e / c
  else:
    sc $\gets$ (b*e - c*d) / D
    tc $\gets$ (a*e - b*d) / D

  # get the difference of the two closest points
  dP $\gets$ w + (sc * u) - (tc * v)
  return norm(dP)


function betaArmSegment(robo):
  # This implements Equations 1 and 2.
  parameters: robo, a Robot
  output: the orientation of the beta arm, a LineSegment

  elbow $\gets$ Vector3()
  elbow.x $\gets$ robo.x + alphaArmLen*cos(robo.alphaCurr)
  elbow.y $\gets$ robo.y + alphaArmLen*sin(robo.alphaCurr)

  fiber $\gets$ Vector3()
  fiber.x $\gets$ elbow.x + betaArmLen*cos(robo.alphaCurr + robo.betaCurr)
  fiber.y $\gets$ elbow.y + betaArmLen*sin(robo.alphaCurr + robo.betaCurr)

  betaSeg $\gets$ LineSegment()
  betaSeg.v0 $\gets$ elbow
  betaSeg.v1 $\gets$ fiber

  return betaSeg


function isCollided(robo):
  parameters: robo, a Robot
  output: True when input Robot is collided with a neighbor, otherwise False

  seg1 $\gets$ betaArmSegment(robo)

  foreach neighbor in robo.neighbors:
    seg2 $\gets$ betaArmSegment(neighbor)
    armSeparation $\gets$ dist3D_Line_to_Line(seg1, seg2)
    if armSeparation <= 2*collisionBuffer + maxDispl:  # Equation 6
        # the two robots are collided exit now
        return True

  # not collided with any neighbor
  return False

function computeCost(robo):
  parameters: robo, a Robot
  output: the cost metric (Equation 8)

  dAlpha $\gets$ robo.alphaCurr - robo.alphaDest
  dBeta $\gets$ robo.betaCurr - robo.betaDest
  return sqrt(dAlpha*dAlpha + dBeta*dBeta)

function computeEnergy(robo):
  parameters: robo, a Robot
  output: the energy metric (Equation 9)

  energy $\gets$ 0
  seg1 $\gets$ betaArmSegment(robo)

  foreach neighbor in robo.neighbors:
    seg2 $\gets$ betaArmSegment(neighbor)
    armSeparation $\gets$ dist3D_Line_to_Line(seg1, seg2)
    energy $\gets$ energy + 1/(armSeparation*armSeparation)

  return energy

function neighborEncroachment(robo):
  parameters: robo, a Robot
  output: True when a neighboring robot is encroaching, otherwise False (Equation 10)

  seg1 $\gets$ betaArmSegment(robo)

  foreach neighbor in robo.neighbors:
    seg2 $\gets$ betaArmSegment(neighbor)
    armSeparation $\gets$ dist3D_Line_to_Line(seg1, seg2)
    if armSeparation <= 2*collisionBuffer + 3*maxDispl:  # Equation 10
        # this neighbor is getting close
        return True

  # no neighbors nearby
  return False


function perturbRobot(robo):
  parameters: robo, a Robot
  output: modify state of input Robot

  # if the positioner is at destination, and no neighbor is encroaching do not do anything
  if computeCost(robo) == 0 and neighborEncroachment(robo) == False:
    return

  # save the current state of the robot's position
  alphaCurr $\gets$ robo.alphaCurr
  betaCurr $\gets$ robo.betaCurr

  # placeholders for minimization search over perturbations
  bestAlpha $\gets$ robo.alphaCurr
  bestBeta $\gets$ robo.betaCurr
  bestScore $\gets$ 1e16 # a large value to be minimized

  # select the metric to minimize (cost or energy)
  minimizeEnergy $\gets$ rand() < phobia

  # randomize order in which perturbations are tried
  shuffle(perturbArray)

  foreach dAlpha, dBeta in perturbArray:
    nextAlpha $\gets$ currAlpha + dAlpha
    nextBeta $\gets$ currBeta + dBeta

    # don't allow out of range moves
    if nextAlpha < 0:
      nextAlpha $\gets$ 0
    if nextAlpha > 360:
      nextAlpha $\gets$ 359.999
    if nextBeta < 0:
      nextBeta $\gets$ 0
    if nextBeta > 360:
      nextBeta $\gets$ 359.999

    # don't overshoot destination coordinates
    if currAlpha > robo.alphaDest and nextAlpha <= robo.alphaDest:
      nextAlpha $\gets$ robo.alphaDest
    if currAlpha < robo.alphaDest and nextAlpha >= robo.alphaDest:
      nextAlpha $\gets$ robo.alphaDest
    if currBeta > robo.betaDest and nextBeta <= robo.betaDest:
      nextBeta $\gets$ robo.betaDest
    if currBeta < robo.betaDest and nextBeta >= robo.betaDest:
      nextBeta $\gets$ robo.betaDest

    # set and test the new position
    robo.alphaCurr $\gets$ nextAlpha
    robo.betaCurr $\gets$ nextBeta

    if isCollided(robo):
      # invalid move, continue to next perturbation
      continue

    if minimizeEnergy:
      # minimize energy metric
      score $\gets$ computeEnergy(robo)
    else:
      # minimize cost metric
      score $\gets$ computeCost(robo)

    if score < bestScore and rand() < greed:
      # new best score found
      bestAlpha $\gets$ nextAlpha
      bestBeta $\gets$ nextBeta
      bestScore $\gets$ score

    else if score == bestScore and rand() < 0.5:
      # score equally good as best seen score
      # take it with a 50 percent chance
      bestAlpha $\gets$ nextAlpha
      bestBeta $\gets$ nextBeta
      bestScore $\gets$ score


  # set the next position for this robot
  robo.alphaCurr $\gets$ bestAlpha
  robo.betaCurr $\gets$ bestBeta
  return

function arrayConverged(robotList):
  parameters: robotList, a list of Robot instances
  output: True when all robots have reached destination, otherwise False

  foreach robo in robotList:
    if computeCost(robo) > 0:
      # this robot has not converged, exit here
      return False

  # all robots have converged
  return True


function recordState(robotList):
  # a function to record the current positions of all Robot
  # instances in the array, called throughout the routine
  # the record the path of each positioner
  parameters: robotList, a list of Robot instances
  output: record the current state the Robot array


# --------------------
# begin main algorithm
# --------------------

# procedure requires a list of each Robot initialized with:
# grid coordinates $(x_\mathrm{b}, y_\mathrm{b})$,
# current coordinates to a set of non-colliding source coordinates,
# and a list of neighboring Robot instances that risk beta arm interference
allRobots $\gets$ list of initialized Robot instances

# record the initial state (all robots at targets)
recordState(allRobots)

# set things in motion
loopIter $\gets$ 1
while loopIter < maxIter:

  foreach robo in allRobots:
    perturbRobot(robo)

  # record new state of robots
  recordState(allRobots)

  # check for convergence
  if arrayConverged(allRobots):
    # all robots have reached destination, exit loop
    break

  loopIter $\gets$ loopIter + 1

# --------------------
# end main algorithm
# --------------------
\end{lstlisting}

\bibliography{bibanticollision}{}
\bibliographystyle{aasjournal}

\end{document}